\documentclass{article}

\usepackage[preprint]{log_2026}

\usepackage{amsfonts}
\usepackage{array}
\usepackage{booktabs}
\usepackage{bm}
\usepackage{colortbl}
\usepackage{enumitem}
\usepackage{graphicx}
\usepackage{longtable}
\usepackage{multirow}
\usepackage{tabularx}
\usepackage[numbers,compress,sort]{natbib}
\usepackage{float}
\usepackage{placeins}

\usepackage{tikz}
\usepackage{pgfplots}

\usetikzlibrary{patterns}
\pgfplotsset{compat=1.18}

\newcommand{\Aerm}{A_{\mathrm{ERM}}}
\newcommand{\Alast}{A_{\mathrm{last}}}
\newcommand{\Ahead}{A_{\mathrm{head}}}
\newcommand{\Afull}{A_{\mathrm{full}}}
\newcommand{\gaincell}[2]{\shortstack{$#1$\\[-1pt]$#2$}}
\newcommand{\uncertaingaincell}[2]{%
  \cellcolor{black!10}\shortstack{$#1$\\[-1pt]$#2$}}

\newcolumntype{L}[1]{>{\raggedright\arraybackslash}p{#1}}

\newif\ifdraftnotes
\draftnotestrue

\newcolumntype{Y}{>{\raggedright\arraybackslash}X}

\title[Disentangling Homophily and Rarity]{Disentangling Homophily and Rarity: Explaining Failure in Graph Neural Networks}

\author[Preben et al.]{
Preben M. Ness
\quad
Fariz Ikhwantri
\quad
Dusica Marijan
\\
Simula Research Laboratory
\\
\texttt{\{prebenmn,fariz,dusica\}@simula.no}
}

\begin{document}

\maketitle

\begin{abstract}
Are heterophilic nodes in a graph harder to classify because they are heterophilic or because they are rare? Some existing work frames classification of such nodes as a subgroup generalisation problem, where a model performs well on the majority group at the expense of the rare group. Others explain this as a problem of neighbourhood aggregation in graph neural networks (GNNs). We assess these two viewpoints through a detailed evaluation of six GNNs on five datasets of varying homophily, and find that homophilic nodes tend to be easier to classify, even when they are rare---challenging the subgroup framing. However, our findings also nuance existing beliefs about how GNNs misrepresent heterophilic nodes. We demonstrate that the information needed to classify heterophilic nodes correctly is often recoverable by retraining the classification head of a model, or even just the final linear classification layer.
\end{abstract}

\section{Introduction}
Graph neural networks (GNNs) often perform unevenly across nodes with different levels of local homophily. On several homophilic graphs, \citet{loveland2023performancediscrepancieslocalhomophily} observe lower performance for nodes which are locally heterophilic. \citet{mao2023demystifyingstructuraldisparitygraph} report the same pattern on heterophilic graphs: locally homophilic nodes can be harder to classify in graphs that are largely heterophilic.

Consequently, one body of existing literature~\citep{NEURIPS2021_08425b88, loveland2023performancediscrepancieslocalhomophily} frames hardness of rare-homophily nodes as a subgroup generalisation problem. Under this explanation, the model learns a spurious correlation~\citep{zhang2026sclgnn} between a node's label and homophily structure---akin to how an image classifier might learn a spurious correlation between the image background and label~\citep{sagawa2020distributionallyrobustneuralnetworks}.

On the other hand, \citet{mao2023demystifyingstructuraldisparitygraph} explain this performance gap as a failure of the GNN's learnt representation of each node. When GNNs aggregate together the feature information of a node and its neighbours, a node's representation becomes increasingly similar to that of its neighbours. Consider the simplified case of binary node classification under different levels of homophily. In a largely heterophilic graph, most nodes of class A will end up looking like their class B neighbours in representation space---and vice versa. A \emph{homophilic} node of class A will then in representation space be indistinguishable from a \emph{heterophilic} node of class B, and be classified incorrectly.

However, even when a classification model fails to classify a sample correctly, we do not know whether that was because information was missing from the sample's representation, or because the classification boundary was drawn poorly. Approaches such as \citet{kirichenko2023lastlayerretraining} demonstrate that the information needed for correct classification is often still present in a model's learnt representation, and performance on rare subgroups can be recovered by cheap retraining of the model's final linear classification layer.

The central theme of our investigation is therefore to shed light on whether rare homophily classification is best understood as a general classification problem, or a phenomenon that arises from the particular way that GNNs process data. We organise our experiments along two overarching questions, which we state here in broad terms, and define precisely in Section~\ref{sec:research-questions}. Firstly, are rare-homophily nodes harder to classify, and does this hold for both rare heterophilic and rare homophilic nodes? Secondly, is the information required to classify a node still recoverable in the GNN's latent representation?

\begin{figure*}[t]
\centering
\begin{minipage}[c]{1.0\textwidth}
    \centering
    \includegraphics[width=\linewidth]
    {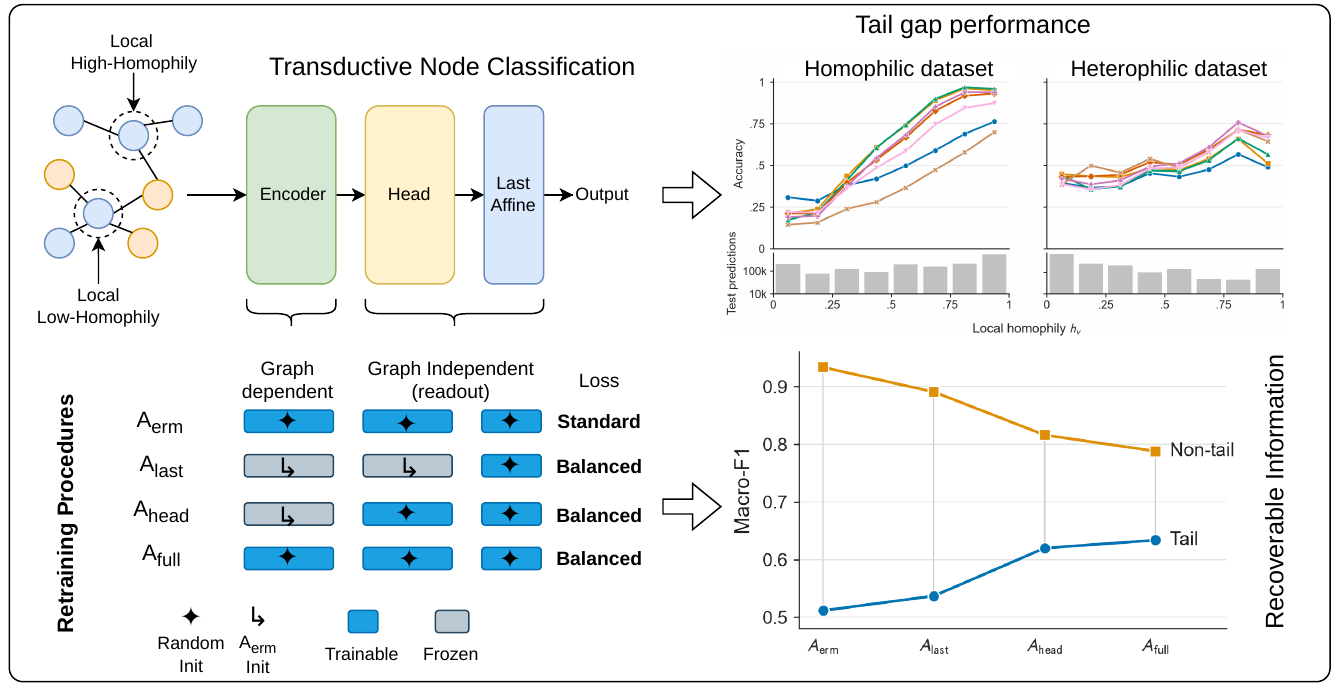}
\end{minipage}
\caption{
    Overview of the diagnostic preparations. The diagram shows the trainable and frozen components, the initialisation source, and the optimisation objective under each preparation.
}
\label{fig:overview}
\end{figure*}

Figure~\ref{fig:overview} illustrates our investigation methods.
Starting from a standard empirical-risk-minimisation (ERM) model, we
reinitialise and retrain the final affine layer, the complete graph-independent
head, or the full model. Full details of sub research questions, estimands, and statistical tests are given in Sections~\ref{sec:definitions} and~\ref{sec:experimental-setup}.

\paragraph{Preregistration and code}
Before performing our experiments and analysis, we preregistered our methodology, RQs, and experimental details. The preregistration can be viewed at the link in this footnote\footnote{Prereg.: \url{https://osf.io/8um5g/files/dxvna?view_only=bc6d73345f57415d9e98e394be2c5407}}, and all deviations are detailed in Appendix~\ref{app:deviations}. Our code is available at the following anonymised link.\footnote{Code: \url{https://anonymous.4open.science/r/homophily-failure-decomp-ED5C}}

\paragraph{Contributions}
In this paper, we conduct a thorough empirical investigation of node classification under varying homophily. Our findings both challenge the framing of this problem as subgroup generalisation and nuance the representation collapse suggested by \citet{mao2023demystifyingstructuraldisparitygraph}.
\begin{itemize}
    \item Firstly, we demonstrate that heterophilic nodes tend to be harder to classify, even when most nodes are heterophilic.
    \item Secondly, we demonstrate that the representations of rare homophily nodes often contain information needed for classification, and that classification head or last layer retraining can recover performance.
\end{itemize}
\section{Related Work}
The challenges of node classification under heterophily, or varying levels of homophily, have been studied broadly and from different angles. Previous studies establish that performance varies with local homophily, but do not separate the effects of local-homophily direction, structural rarity, and class composition. Existing heterophily methods train new graph representations to improve performance, but do not determine whether the original representation already supports a better classifier. Existing refitting studies show that poor subgroup performance need not imply a deficient representation, but this distinction has not been examined for local-homophily disparities in GNNs.

\paragraph{Graph Homophily}
Prior work shows that graph neural network (GNN) performance varies with local homophily, but uses different definitions of the relevant node groups. \citet{loveland2023performancediscrepancieslocalhomophily} study nodes whose local homophily differs from the graph-wide pattern, while \citet{mao2023demystifyingstructuraldisparitygraph} compare common and uncommon neighbourhood patterns and measure their similarity to labelled training nodes. \citet{du2022gbkgnngatedbikernelgraph} report performance across fixed local-homophily intervals, and \citet{bi2022makeheterophilygraphsbetter} show that experimentally changing local homophily changes classification accuracy. These studies do not determine whether a node group is difficult because its homophily is low, because its neighbourhood pattern is uncommon, or because it contains different classes.

Research on subgroup generalisation and distribution shift asks whether models fail on node types that are poorly represented among training nodes. \citet{NEURIPS2021_08425b88} group test nodes by their distance from the training data in aggregated-feature space and find that accuracy declines with distance. In synthetic graphs, \citet{loveland2023performancediscrepancieslocalhomophily} find smaller disparities when labelled nodes represent local-homophily patterns more uniformly. \citet{mao2023demystifyingstructuraldisparitygraph} train on the common neighbourhood pattern and find poor transfer to the uncommon pattern. \citet{loveland2024unveilingimpactlocalhomophily} and \citet{yang2025leveraginginvariantprincipleheterophilic} instead impose differences between training and test local-homophily distributions. \citet{yang2025leveraginginvariantprincipleheterophilic} also compare equally sized high- and low-homophily test groups. Taken together, these studies suggest that performance may depend both on a node's local homophily and on how common its neighbourhood pattern is. The respective roles of these factors, and of class composition, remain difficult to separate. We aim to disentangle them by comparing rare low- and high-homophily tails, adjusting for class composition, and performing targeted graph interventions.

\paragraph{Heterophily-designed Architectures}
A multitude of architectures have been proposed to handle the problem of heterophilic node classification. While these models are often motivated by assumptions about heterophilic failure, their goal is to improve predictive performance, not necessarily to understand the underlying phenomenon.

Existing approaches address heterophily by changing how graph information is processed. H2GCN~\citep{zhu2020beyondhomophilygnns} preserves ego and multi-hop representations separately, GPR-GNN~\citep{chien2021adaptiveuniversalgeneralizedpagerank} learns graph-wide weights over propagation depths, and ACM-GCN~\citep{luan2022revisitingheterophilygnns} and Node-MoE~\citep{han2025nodewisefilteringgraphneural} choose among different graph filters for each node. LINKX~\citep{lim2021largescalelearningnonhomophilous} avoids neighbourhood feature propagation altogether, while Mowst~\citep{zeng2024mixtureweakstrong} routes nodes between feature-only and graph-dependent predictors. Other methods rewire the graph~\citep{bi2022makeheterophilygraphsbetter} or train an invariant encoder~\citep{yang2025leveraginginvariantprincipleheterophilic}. These interventions show that redesigned models can improve accuracy, but they change the representation and classifier together. They therefore do not reveal which component produced the disparity in the original model.

\paragraph{Assessing representations}
Broadly speaking, a model's predictions depend on the class information available in its representation and on the classifier fitted to that representation. Probing separates these contributions by freezing the representation and fitting a new classifier above it. Linear and non-linear probes implement the same diagnostic using classifiers of different capacity~\citep{alain2016understanding,belinkov-2022-probing,pmlr-v206-akhondzadeh23a}. Balanced classifier refitting has shown that frozen representations can support substantially better predictions for rare classes and poorly performing groups than the model's original classifier produces~\citep{kang2019decoupling,kirichenko2023lastlayerretraining}. To our knowledge, frozen-representation refitting has not been used to analyse performance differences between local-homophily groups. We address this gap by comparing refitting of the final linear layer and the complete graph-independent head with full-model retraining. This comparison shows how much performance can be recovered while preserving the graph-dependent representation, and what additional recovery follows when that representation is relearned.
\section{Definitions and Research Questions}\label{sec:definitions}

\paragraph{Transductive node classification}
Let \(G = (V, E)\) be an undirected graph with \(n = |V|\) nodes. Each node \(v \in V\) has a feature vector \(\bm{x}_v \in \mathbb{R}^d\) and a class label \(y_v \in \{1, \ldots, C\}\). The set of nodes is partitioned into training, validation, and test splits. The graph and features are observed for all nodes, and the model trains using the labels of the training split and is evaluated on the labels of the test split. The validation set is used for checkpoint selection (see further training details in Section~\ref{sec:experimental-setup} and Appendix~\ref{app:model-settings}).

\paragraph{Local homophily}\label{sec:local-homophily}

For a node \(v\), local homophily \(h_v\) is defined as
\begin{equation}
    h_v =
    \frac{
        \left| \{ u \in \mathcal{N}(v):y_u=y_v \} \right|}
    {
        |\mathcal{N}(v)|
    },
\end{equation}
where \(\mathcal{N}(v)\) is the one-hop neighbourhood of \(v\)
\citep[Definition~2]{loveland2023performancediscrepancieslocalhomophily}.
Thus, \(h_v=1\) when every neighbour shares \(v\)'s label and \(h_v=0\)
when none do.\footnote{This way of defining \(h_v\) is not applicable to isolated nodes, but none of the datasets we used contain isolated nodes after standard preprocessing.}

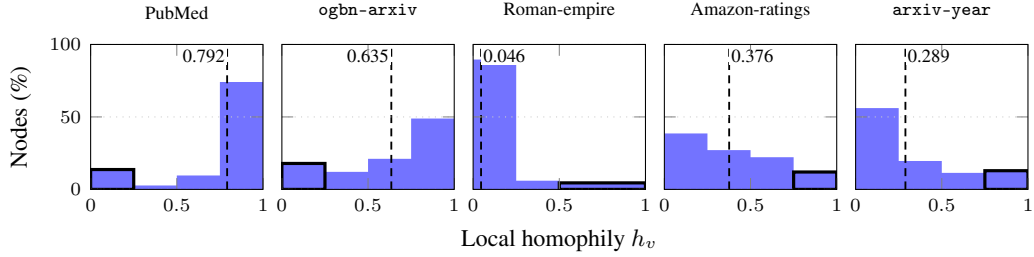
\begin{figure}[t]
\centering
\usepgfplotslibrary{groupplots}
\begin{tikzpicture}
\begin{groupplot}[
    group style={
        group size=5 by 1,
        horizontal sep=0.25cm,
        ylabels at=edge left,
        yticklabels at=edge left
    },
    scale only axis,
    width=2.28cm,
    height=1.92cm,
    xmin=0, xmax=1,
    ymin=0, ymax=100,
    xtick={0,0.5,1},
    ytick={0,50,100},
    tick label style={font=\scriptsize},
    title style={font=\scriptsize},
    label style={font=\footnotesize},
    ymajorgrids=true,
    grid style={dotted,gray!40},
    axis on top,
]

\nextgroupplot[
    title={PubMed},
    ylabel={Nodes (\%)}
]
\addplot+[ybar interval, fill=blue!55, draw=none, mark=none]
coordinates {
    (0,13.7)
    (0.25,2.7)
    (0.5,9.6)
    (0.75,74.1)
    (1,0)
};
\addplot+[ybar interval, fill=blue!55, draw=black, line width=1.2pt, mark=none]
coordinates {(0,13.7) (0.25,0)};
\draw[densely dashed,black,line width=0.7pt]
    (axis cs:0.792,0) -- (axis cs:0.792,100);
\node[font=\scriptsize,anchor=north east,fill=white,inner sep=0.7pt]
    at (axis cs:0.792,98) {0.792};

\nextgroupplot[title={\texttt{ogbn-arxiv}}]
\addplot+[ybar interval, fill=blue!55, draw=none, mark=none]
coordinates {
    (0,17.9)
    (0.25,12.1)
    (0.5,21.1)
    (0.75,48.9)
    (1,0)
};
\addplot+[ybar interval, fill=blue!55, draw=black, line width=1.2pt, mark=none]
coordinates {(0,17.9) (0.25,0)};
\draw[densely dashed,black,line width=0.7pt]
    (axis cs:0.635,0) -- (axis cs:0.635,100);
\node[font=\scriptsize,anchor=north east,fill=white,inner sep=0.7pt]
    at (axis cs:0.635,98) {0.635};

\nextgroupplot[
    title={Roman-empire},
    xlabel={Local homophily \(h_v\)}
]
\addplot+[ybar interval, fill=blue!55, draw=none, mark=none]
coordinates {
    (0,89.6)
    (0.25,6.0)
    (0.5,4.5)
    (1,0)
};
\addplot+[ybar interval, fill=blue!55, draw=black, line width=1.2pt, mark=none]
coordinates {(0.5,4.5) (1,0)};
\draw[densely dashed,black,line width=0.7pt]
    (axis cs:0.046,0) -- (axis cs:0.046,100);
\node[font=\scriptsize,anchor=north west,fill=white,inner sep=0.7pt]
    at (axis cs:0.046,98) {0.046};

\nextgroupplot[title={Amazon-ratings}]
\addplot+[ybar interval, fill=blue!55, draw=none, mark=none]
coordinates {
    (0,38.6)
    (0.25,27.1)
    (0.5,22.2)
    (0.75,12.0)
    (1,0)
};
\addplot+[ybar interval, fill=blue!55, draw=black, line width=1.2pt, mark=none]
coordinates {(0.75,12.0) (1,0)};
\draw[densely dashed,black,line width=0.7pt]
    (axis cs:0.376,0) -- (axis cs:0.376,100);
\node[font=\scriptsize,anchor=north west,fill=white,inner sep=0.7pt]
    at (axis cs:0.376,98) {0.376};

\nextgroupplot[title={\texttt{arxiv-year}}]
\addplot+[ybar interval, fill=blue!55, draw=none, mark=none]
coordinates {
    (0,56.1)
    (0.25,19.6)
    (0.5,11.4)
    (0.75,12.9)
    (1,0)
};
\addplot+[ybar interval, fill=blue!55, draw=black, line width=1.2pt, mark=none]
coordinates {(0.75,12.9) (1,0)};
\draw[densely dashed,black,line width=0.7pt]
    (axis cs:0.289,0) -- (axis cs:0.289,100);
\node[font=\scriptsize,anchor=north west,fill=white,inner sep=0.7pt]
    at (axis cs:0.289,98) {0.289};

\end{groupplot}
\end{tikzpicture}
\caption{Distribution of nodes across local-homophily bins. Bar
widths show the bin intervals and heights show within-dataset percentages.
Dashed lines give mean \(h_v\); thick outlines mark the selected tails.}
\label{fig:local-homophily-bins-tikz}
\end{figure}

\paragraph{Local-homophily bins}
We initially assign all nodes in a graph to four equal-width $h_v$ bins, similarly to \cite{loveland2023performancediscrepancieslocalhomophily}, and merge any bin containing fewer than $1\%$ of a graph's total nodes with an adjacent bin. See Appendix~\ref{app:local-homophily-bins} for further details and the exact construction algorithm. The resulting bins are shown in Figure~\ref{fig:local-homophily-bins-tikz}. Note that the bin assignments are determined only by the dataset and are identical across splits, seeds, models, and preparations.

\paragraph{Nomenclature}
We call a dataset homophilic when the mean \(h_v\) is above \(0.5\), and heterophilic otherwise. We refer to nodes whose \(h_v\) value is uncommon in a dataset as \emph{rare homophily} nodes, and the extreme \(h_v\) bin in each dataset as the \emph{rare homophily tail}. This tail is therefore \textbf{the lowest-\(h_v\) bin in a homophilic dataset and the highest-\(h_v\) bin on a heterophilic dataset}.

\paragraph{Retraining procedures}\label{sec:preps}

We treat each model as a graph-dependent encoder \(h\) followed by a non-linear head \(g\) and a final linear classifier \(f\). For node \(v\) we then have

\begin{equation}
    \bm{y}^{\text{pred}}_v = f(\bm{z}_v) = \bm{W}_f\bm{z}_v + \bm{b}_f, \qquad
    \bm{z}_v = g(\hat{\bm{z}}_v), \qquad
    \hat{\bm{z}}_v = h(G, \bm{x}_v)
\end{equation}

where \(\bm{y}^{\text{pred}}_v\) is the vector of logits for node \(v\), and \(\bm{W_f}\) and \(\bm{b}_f\) are weights matrix and bias vector respectively of $f$.

For each model architecture, we discard the final prediction layer and take the remaining trunk as our encoder \(h\) (see Appendix \ref{app:model-settings} for exactly where we cut each model). We then attach our head \(g\) and linear classifier \(f\), with the following shared architecture:

\begin{equation}
    g(\hat{\bm{z}}_v) =
    \operatorname{Dropout}_{0.2}\!\left[
        \operatorname{GELU}\!\left(
            \operatorname{LayerNorm}(
                \bm{W}_g\hat{\bm{z}}_v + \bm{b}_g
            )
        \right)
    \right],    
    \qquad
    f(\bm{z}_v) = \bm{W}_f\bm{z}_v + \bm{b}_f
\end{equation}

with hidden widths of 512. The encoder output is also 512-dimensional, except
for the model H2GCN-2, where it is 511-dimensional.\footnote{The architecture of H2GCN-2 requires the output dimension to be a multiple of 7.}

As shown in Figure \ref{fig:overview}, we start by training each model using average empirical risk minimisation (ERM) over all nodes. This is the standard training procedure, and we call it \(\Aerm\). We then perform three retraining procedures on the model, this time with a balanced GroupDRO loss \citep{sagawa2020distributionallyrobustneuralnetworks,sagawa2020groupdrocode}, using the homophily bins as the GroupDRO groups. In \(\Alast\) we retrain the final layer $f$, in \(\Ahead\) we retrain the whole classification head \(f \circ g\), and in \(\Afull\) we retrain the whole model from scratch.

\subsection{Research questions}\label{sec:research-questions}

With the above definitions and framework, we concretise our RQs as follows:

\begin{description}
    \item[\textbf{RQ1}]
    Do trained models perform worse on the rare homophily tail than on the test set as a whole, and does this hold for both homophilic and heterophilic datasets?
    \begin{description}
        \item[RQ1a]
        Does the gap persist within classes: do tail nodes score below non-tail nodes of the same class?
        \item[RQ1b]
        Do the aggregation-based models (GCN, GraphSAGE-mean, H2GCN-2, GPR-GNN, ACM-GCN) show larger tail gaps than the graphless MLP on the same dataset?
        \item[RQ1c]
        Do the heterophily-designed aggregation models (H2GCN-2, GPR-GNN, ACM-GCN) show smaller tail gaps than the standard baselines (GCN and GraphSAGE-mean)? 
    \end{description}
    \item[\textbf{RQ2}]
    When a tail gap exists, where is recoverable information lost?
    \begin{description}
        \item[RQ2a]
        In the final linear layer \(f\)?
        \item[RQ2b]
        In the graph-independent classifier \(f \circ g\)?
        \item[RQ2c]
        In the graph-dependent trunk \(h\)?
    \end{description}
\end{description}

\section{Experimental Setup}\label{sec:experimental-setup}

\paragraph{Datasets}\label{sec:dataset}
We use five datasets in total, two homophilic and three heterophilic under the definition in Section~\ref{sec:local-homophily}. The two homophilic are PubMed \citep{yang2016revisitingsemisupervisedlearninggraph} and \texttt{ogbn-arxiv}~\citep{hu2021opengraphbenchmarkdatasetsmachine}, and the three heterophilic are Roman-empire~\citep{platonov2024criticallookevaluationgnns}, Amazon-ratings~\citep{platonov2024criticallookevaluationgnns}, and \texttt{arxiv-year}~\citep{lim2021largescalelearningnonhomophilous}. Note that \texttt{arxiv-year} and \texttt{ogbn-arxiv} are the same graph, but with different labels.

For a given model, dataset, and training procedure, we perform 30 training runs. These 30 runs correspond to different combinations of dataset splits and random model training seeds. PubMed uses the ten Geom-GCN splits \citep{pei2020geomgcn} with seeds 1--3. Roman-empire and Amazon-ratings use the ten released 50/25/25 Platonov splits \cite{platonov2024criticallookevaluationgnns} with seeds 1--3. \texttt{ogbn-arxiv} uses the official OGB time split with seeds 1--30. \texttt{arxiv-year} uses the five 50/25/25 splits from \citet{lim2021largescalelearningnonhomophilous} with seeds 1--6.

\paragraph{Models and training}\label{sec:model-training} We use seven models in total, six GNNs and a graphless MLP baseline. For GNNs, we use the standard aggregation models GCN~\citep{kipf2017semisupervisedclassificationgraph} and GraphSAGE-mean~\citep{hamilton2018inductiverepresentationlearninglarge}, and the heterophily-designed aggregation models H2GCN-2~\citep{zhu2020beyondhomophilygnns}, GPR-GNN~\citep{chien2021adaptiveuniversalgeneralizedpagerank}, and ACM-GCN~\citep{luan2022revisitingheterophilygnns}. Additionally, we use the non-aggregation model LINKX~\citep{lim2021largescalelearningnonhomophilous}.

All runs train for their complete training budget. ERM checkpoints are selected by validation accuracy. The three GroupDRO procedures select checkpoints by validation macro-F1 on the rare homophily tail. \(\Afull\) uses the ERM optimiser, learning rate, schedule, and budget for its model. \(\Alast\) and \(\Ahead\) use the same optimiser at one tenth of the ERM learning rate and train for ten times the ERM budget. Appendix~\ref{app:model-settings} gives model-specific training settings and further experimental details.

\subsection{Outcomes and estimands}\label{sec:stats}
For individual homophily bins, we typically use macro-F1 as the performance metric, following \citet{loveland2023performancediscrepancieslocalhomophily}. In a few cases, particularly for the class-imbalanced dataset Roman-empire, not all bins contain nodes of each class. In those cases, the F1 values for those classes are undefined. We therefore compute macro-F1 only over the classes represented in the bin. This avoids assigning arbitrary scores to classes that are absent and ensures that a perfect predictor scores \(1.00\). Consequently, different bins may average over different sets of classes.

\paragraph{Class adjustment}
Since a difference in performance between the rare homophily tail and the test set as a whole could be explained by the tail having a more challenging class composition, we specify an operation for class composition adjustment as follows. Let \(C\) denote the set of all classes that are present in the tail bin. Construct the confusion matrix \(M_{\text{tail}}\) over classes in \(C\) for the tail bin. To obtain a class-adjusted tail score, reweight the rows of \(M_{\text{tail}}\), where the weight is the proportion of that row's true class in the whole test set. We can then compute metrics, such as macro-F1 or accuracy, from the reweighted confusion matrix. The exact implementation details of this adjustment procedure are given in Appendix~\ref{app:rq2-within-class}.

\paragraph{The tail gap}
For a run \(r\), we define the \emph{tail gap} \(G_r^m\) of a performance metric \(m\) as the value of \(m\) over the whole test set minus the value over the tail bin. For example, the macro-F1 tail gap \(G_r^{\text{macro-F1}} = \operatorname{macro-F1}_{\text{test}} - \operatorname{macro-F1}_{\text{tail}}\). Unless otherwise noted, these are unadjusted for class composition. The adjusted tail gap is then the gap value after the class composition adjustment described above. We compute performance metrics, tail gaps, and differences after retraining procedures for each of the 30 runs and then report their means and 95\% confidence intervals.

\paragraph{Uncertainty and statistical significance}
For every reported tail gap, or retraining gain, we resample the 30 matched run-level values with replacement 1{,}000 times and report the 2.5th and 97.5th percentiles of the bootstrap means \citep{efron1994introduction,davison1997bootstrap}. The quantiles use linear interpolation and are computed without multiplicity correction. We treat the 30 dataset split/random seed runs as our independent resampling units.

Since we measure differences between quantities---tail performance vs whole test, change after retraining---we claim statistical significance when the 95\% confidence interval of the measured difference excludes 0. That is, we claim that the measured quantity is statistically significantly positive or negative.

For RQ2, where we look at the share of performance that can be recovered by retraining different parts of a model, we define for \(p \in \{\mathrm{last}, \mathrm{head}\}\), the \emph{recovered share} as

\[
    \rho_p =
    \frac{\hat A_p-\hat A_{\mathrm{erm}}}
         {\hat A_{\mathrm{full}}-\hat A_{\mathrm{erm}}},
\]
where \(\hat{A}_p\) is the macro-F1 score over the tail after retraining procedure $p$, averaged over the 30 training runs. For numerical stability, we interpret \(\rho_p\) only when the denominator is at least 0.03, and claim statistical significance only when the 95\% confidence interval of the numerator excludes zero.
\section{Results and Analysis}
\subsection{RQ1: Is there a tail gap?}
\begin{figure}[t]
  \centering
  \includegraphics[width=\linewidth]{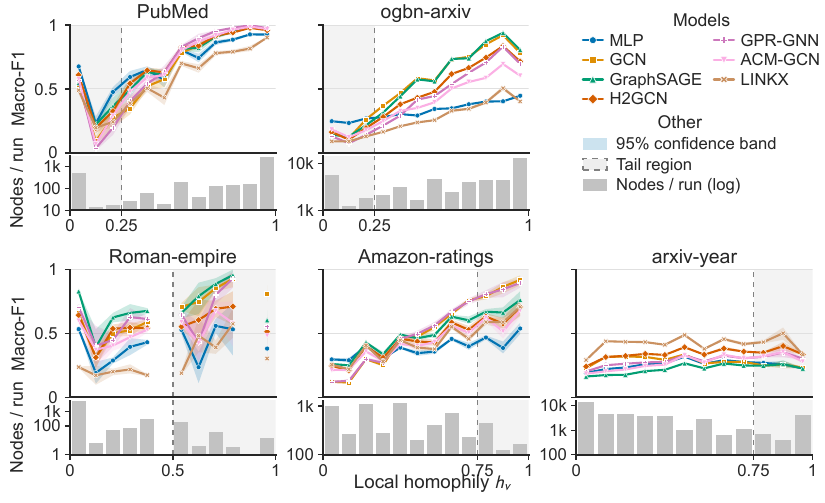}
  \caption{
    \textbf{RQ1}. Average test macro-F1 across 12 equal-width local-homophily bins. Grey bars show the mean test-node count per run for each bin, and grey regions mark each dataset's rare homophily tail. Roman-empire macro-F1 is omitted where a bin lacks test nodes in at least one split.
    }
  \label{fig:rq1-continuous-macro-f1}
\end{figure}

Figure~\ref{fig:rq1-continuous-macro-f1} plots average prediction macro-F1 as a function of \(h_v\), for 12 equal-width \(h_v\) bins. Because of its skewed distribution over \(h_v\) bins, the plot for Roman-empire is noisy and difficult to interpret. However, among the remaining four datasets, the plots indicate a general trend where models score better on average on locally homophilic nodes.

\begin{figure}[t]
  \centering
  \includegraphics[width=\linewidth]
  {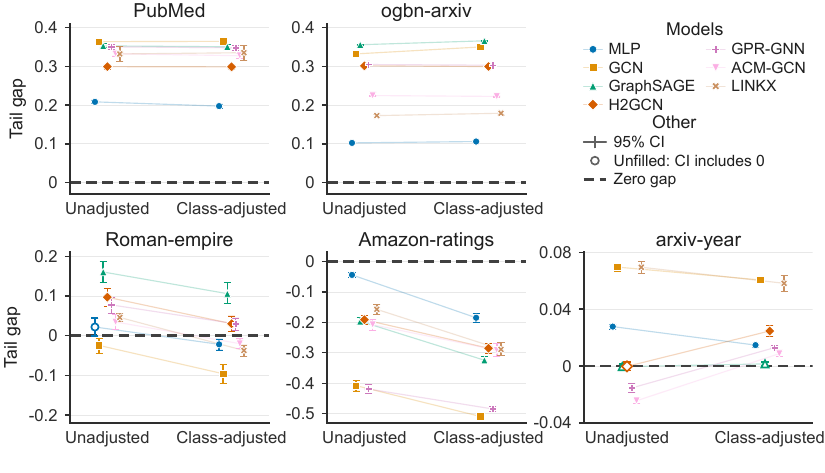}
  \caption{
    \textbf{RQ1, RQ1a}. Macro-F1 tail gaps for all models on all datasets, with homophilic datasets on the top and heterophilic datasets on the bottom. Unadjusted and class-adjusted values are shown on the left and right, respectively, of each subplot, and lines connect values for a given model. Vertical lines denote 95\% confidence intervals, and models whose interval contains zero are denoted with an unfilled marker symbol.
    }
  \label{fig:rq1-rq2-tail-gaps}
\end{figure}

Figure~\ref{fig:rq1-rq2-tail-gaps} plots the unadjusted and class-adjusted macro-F1 tail gap for all models and all datasets. Both homophilic datasets show a positive tail gap for all models, meaning that tail performance is worse than average test performance. The three heterophilic datasets are more varied. Amazon-ratings shows a negative tail gap for all models, meaning that performance was \emph{better} on the rare homophilic tail. Roman-empire initially shows a positive tail gap for most models, but this pattern partially reverses when adjusted for class composition. \texttt{arxiv-year} is the heterophilic dataset that shows the clearest positive tail gap, although the size is small (c.f. Figure~\ref{fig:rq5-preparations-overview}). The exact unadjusted means and confidence intervals are given in Table~\ref{tab:erm-tail-gap} in Appendix~\ref{app:rq1}, and the corresponding adjusted figures in Table~\ref{tab:rq2-adjusted-macro-f1} in Appendix~\ref{app:rq2-within-class}.

\paragraph{RQ1a: Is the tail gap a class composition artefact?}
The tail gaps for the two homophilic datasets are essentially unchanged when adjusting for class composition. The class adjustment for Roman-empire and Amazon-ratings reduces the value of all gaps, meaning that tail difficulty is at least partially explained by the tail containing disproportionately more of the difficult classes.

Appendix~\ref{app:edge-perturbation} shows the results of an experiment where we progressively randomly changed edges in a graph to either i) move \(h_v\) towards the graph mean or ii) randomly, in both cases preserving node degree. We then evaluated the 30 ERM-trained final checkpoints of the models GCN and GraphSAGE-mean on the rewired graphs. In all cases, we find that making nodes more homophilic increases model performance, even when homophilic nodes are rare.

\begin{figure}[ht]
  \centering
  \includegraphics[width=\linewidth]
  {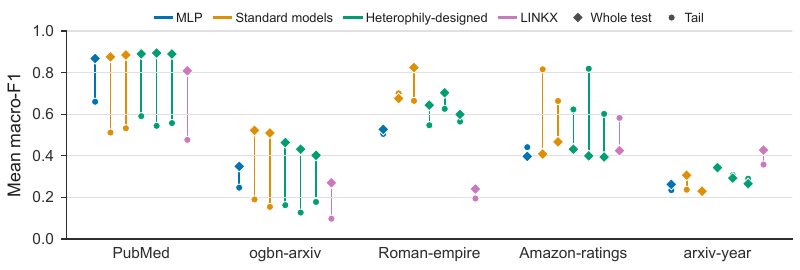}
  \caption{
    \textbf{RQ1b, RQ1c}. Whole-test and tail macro-F1 for all models on all datasets, with models coloured by model family. Within each dataset, models are ordered from left to right as MLP, GCN, GraphSAGE-mean, H2GCN-2, GPR-GNN, ACM-GCN, and LINKX. Blue denotes MLP, orange the standard aggregation models, green the heterophily-designed aggregation models, and purple denotes LINKX.
    }
  \label{fig:rq3-rq4-whole-tail-scores}
\end{figure}

\paragraph{RQ1b: Do aggregation models have larger gaps than MLP?}
Figure~\ref{fig:rq3-rq4-whole-tail-scores} shows the whole test and tail macro-F1 scores for all models on all datasets, showing each tail gap. Comparing the aggregation-based models (orange and green) with the MLP (blue), we see that the MLP has the smaller gap on both the homophilic datasets and outperforms all other models on the tail bin. On all three heterophilic datasets, the MLP tail gap is small in magnitude, with the gaps of the other models varying. Appendix~\ref{app:rq3-aggregation-vs-mlp} gives the exact plotted values and their confidence intervals.

\paragraph{RQ1c: Do heterophily-designed aggregation models have smaller tail gaps?}
Comparing the heterophily-designed aggregation models (green) and the standard aggregation models (yellow) in Figure~\ref{fig:rq3-rq4-whole-tail-scores}, we note that the heterophily designs show a clear tendency for slightly smaller tail gaps on the homophilic datasets. But we also note that the heterophilic designs do not show the same pattern for the heterophilic datasets. Neither do these models perform clearly better than the standard baseline models or even the MLP model on the heterophilic datasets. Appendix~\ref{app:rq4-designed-vs-standard} gives the exact plotted values and their confidence intervals.

\subsection{RQ2: Where is information lost?}
\begin{figure}[t]
  \centering
  \includegraphics[width=\linewidth]{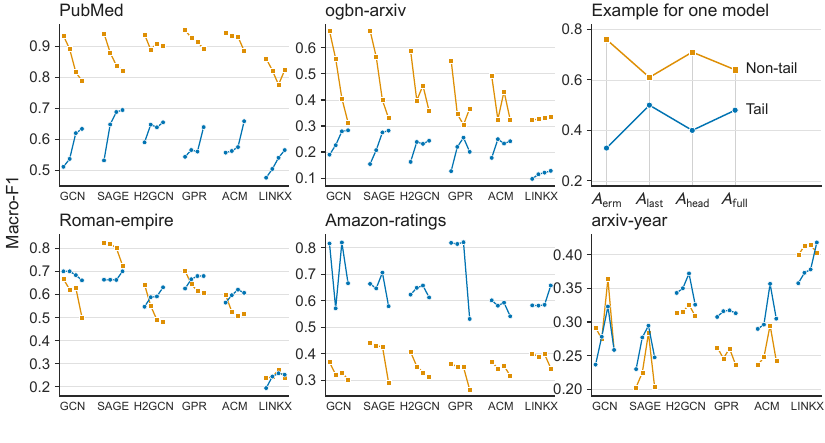}
  \caption{
        \textbf{RQ2, RQ2a, RQ2b, RQ2c}. Mean tail and non-tail macro-F1 for different training procedures across the six graph models. Within each model, points from left to right are \(\Aerm\), \(\Alast\) (RQ2a), \(\Ahead\) (RQ2b), and \(\Afull\) (RQ2c). Circles show tail scores and squares show non-tail scores. Note that each sub-figure uses its own vertical scale. An increase in the blue curve indicates recovered performance on the rare-homophily tail, while a decrease in the orange curve indicates corresponding performance degradation on the rest of the test set.
    }
  \label{fig:rq5-preparations-overview}
\end{figure}

Figure~\ref{fig:rq5-preparations-overview} compares tail and non-tail macro-F1 across the procedures. The extent to which retraining helps improve tail performance varies between models and datasets. Among the 17 model/dataset cells with statistically significant results, final-layer retraining (RQ2a) exceeds 70\% in eight, and head-retraining exceeds 70\% in 13. Although the homophilic datasets show more consistent recovery than the heterophilic ones, we observe that frozen representations often contain information needed for correct node classification. The performance recovered by retraining depends on the dataset and model, and recovery. There is a consistent trade-off, where increased tail performance often is at the expense of degraded performance on the rest of the test set. Appendix~\ref{app:rq5-preparation-scores} reports the cell-level gains and confidence intervals.

\paragraph{RQ2: Sensitivity to retraining method}
We use GroupDRO as our balanced retraining loss in RQ2. This is a standard approach, but one of many possible options. As a sensitivity test, we retrained a subset of models on PubMed using DFR-style last-layer logistic refitting adapted from \citet{kirichenko2023lastlayerretraining}, and FG-CCDB group-balanced training adapted from \citet{zhao2026finegrainedccdb}. DFR was competitive with GroupDRO in some cases, and FG-CCDB failed to retrain well in all tested cases. Neither set of results invalidates our use of GroupDRO. The detailed results are given in Appendix~\ref{app:rq5-sensitivity-retraining}.
\section{Conclusion}
Our findings support the established notion that locally heterophilic nodes in homophilic graphs are harder to classify. But we do not find clear evidence that the effect holds in the opposite direction. Our findings are about equally well explained by locally heterophilic nodes being harder, regardless of whether they are rare or common in a given graph. Therefore, our results do not clearly support prior findings~\citep{loveland2023performancediscrepancieslocalhomophily, mao2023demystifyingstructuraldisparitygraph} that locally homophilic nodes can be difficult to classify in heterophilic graphs. Much of the apparent hardness of rare homophilic nodes is explained by class composition.

Although the extent varies between models and datasets, we find that frozen model representations often contain the required information needed to classify rare-homophily nodes correctly. Balanced retraining of the classifier head and even just the last linear layer often improves tail performance substantially. Our findings at least nuance the existing framing of the extent to which GNN representation aggregation damages the information needed for correct classification of rare-homophily nodes~\cite{mao2023demystifyingstructuraldisparitygraph}.

\paragraph{Limitations and future work}
The homophily bins are oracle quantities. Computing \(h_v\) uses true labels from the full graph, and a training node's bin can depend on its neighbours' held-out labels. An interesting avenue of future work would be to use estimated \(h_v\) values, without knowledge of validation or test-set labels.

There is no single definite way to measure the information contained in a given latent representation~\citep{belinkov-2022-probing, pmlr-v206-akhondzadeh23a}. We use retraining procedures to assess if the frozen representation can be used to successfully classify rare-homophily nodes. This measurement is flexible but one-directional. If the retraining succeeds, we know that the information is there, but if it does not, we cannot say for sure whether the information is absent or the retraining failed for some other reason.

We trained seven models on five transductive benchmark datasets, with each configuration evaluated across 30 split--seed combinations. The statistical bootstrap tests then measure variation over splits and seeds only. Expanding the experiments to include more graphs, models, training settings, and sources of training variation would tell us more about how our findings generalise.

Future work could examine whether graph foundation models~\citep{10.1145/3543507.3583386, 10.1145/3696410.3714828} pretrained across graphs with different homophily regimes exhibit similar performance disparities on rare local-homophily patterns. The subgroup definitions and
component-wise retraining procedures developed here provide a starting point for auditing recoverability during downstream adaptation.

The main results presented in this paper are descriptions of observed effects, and further study is needed to understand and describe the underlying mechanisms. Our exploratory edge interventions in Appendix~\ref{app:edge-perturbation} provide preliminary evidence that neighbourhood label composition contributes to tail-node performance. But disentangling the effects of difficulty variations between nodes, node degree, class composition, and other confounding variables from the effect of homophily will require a range of carefully designed and targeted future experiments.

\clearpage
\appendix
\raggedbottom

\section{Deviations from Preregistration}
\label{app:deviations}
\noindent The full analysis protocol, methodology, and experimental settings were registered on 10 July 2026 and can be inspected at the following anonymised link:

\url{https://osf.io/8um5g/files/dxvna?view_only=bc6d73345f57415d9e98e394be2c5407}.

Table~\ref{tab:deviations} records all deviations.

\begin{table}[h]
  \centering
  \caption{Ledger of deviations from the preregistered analysis and experimental protocol.}
  \label{tab:deviations}
  \begin{tabularx}{\linewidth}{@{}Y@{}}
    \toprule
    Deviation \\
    \midrule
    The registered code used scikit-learn's default label set for macro-F1. This takes the union of the labels in the true values and predictions, so the classes included in the average could vary between evaluations. We replaced this with the classes present in the evaluated subset's true labels. The Roman-empire recovery experiments were rerun because the error also affected the validation score used for checkpoint selection. All macro-F1 scores were then recalculated using the corrected definition.\\
    \midrule
    The preregistration described a rule where only classes with at least 10 tail test nodes should be included in class-wise accuracy comparisons. In contrast, for macro-F1, all classes represented among the tail nodes. Although motivated by removing noisy results, we felt the rule was too arbitrary and changed it so that class-wise accuracy now uses the same rule as the existing macro-F1. All dependent estimates, tables, and figures were recalculated. The only visible difference on the plotted figures was that the class-adjustment of Roman-empire in Figure~\ref{fig:rq1-rq2-tail-gaps} became less conclusive, with the adjusted models landing either side of 0.\\
    \midrule
    Minor grammatical, nomenclature, and notation adjustments were made to the RQ formulations. For example, we use the term \emph{rare homophily tail} in this paper, instead of \emph{minority pattern tail}, which is used in the preregistration.\\
    \bottomrule
  \end{tabularx}
\end{table}

\section{Details of Model Architectures and Training Settings}
\label{app:model-settings}

\paragraph{Shared Platonov backbone.}
MLP, GCN, and GraphSAGE-mean use the five-block residual architecture of
\citet{platonov2024criticallookevaluationgnns}. With dropout 0.2 throughout,
\[
\begin{aligned}
h^{(0)}&=\operatorname{GELU}
  \bigl(\operatorname{Dropout}(W_{\mathrm{in}}x+b_{\mathrm{in}})\bigr),
  &W_{\mathrm{in}}&:\dim(x)\to512,\\
h^{(l)}&=h^{(l-1)}+
  \operatorname{Block}_l\bigl(\operatorname{LayerNorm}(h^{(l-1)})\bigr),
  &l&=1,\ldots,5,\\
z&=\operatorname{LayerNorm}(h^{(5)}).
\end{aligned}
\]
Each model-specific operation in Table~\ref{tab:models} is followed by
\[
\operatorname{FFN}(x)=
\operatorname{Dropout}\!\left(
W_2\operatorname{GELU}\!\left(
\operatorname{Dropout}(W_1x+b_1)\right)+b_2\right),
\qquad W_2:512\to512,
\]
where \(W_1\) maps the model-specific operation's output to width 512.

\begin{table}[H]
\centering
\begin{tabular}{@{}L{0.14\textwidth}L{0.50\textwidth}L{0.28\textwidth}@{}}
\toprule
Model & Encoder & Training \\
\midrule
MLP~\citep{platonov2024criticallookevaluationgnns} & Shared backbone; model-specific operation is the identity, so each block has no graph access. & AdamW, lr $3{\cdot}10^{-5}$, wd 0, 1000 steps. \\
GCN~\citep{kipf2017semisupervisedclassificationgraph} & Shared backbone; parameter-free symmetric degree-normalised neighbour aggregation ($1/\sqrt{d_u d_v}$). Self-loops added. & AdamW, lr $3{\cdot}10^{-5}$, wd 0, 1000 steps. \\
GraphSAGE-mean~\citep{hamilton2018inductiverepresentationlearninglarge} & Shared backbone; concatenation of the node state and mean neighbour state (1024-dimensional), followed by an FFN with widths $1024{\to}512{\to}512$. No self-loops. & AdamW, lr $3{\cdot}10^{-5}$, wd 0, 1000 steps. \\
H2GCN-2~\citep{zhu2020beyondhomophilygnns} & Ego and neighbour separation over exact 1-hop and 2-hop indicator matrices, each symmetrically normalised; \texttt{network\_setup = M73-R-T1-G-V-T2-G-V-C1-C2-D0.5}; internal width 73; input-feature normalisation on; dropout 0.5. $z_v =$ the native 7-block concatenation (511-dim); the source output layer is removed. & Adam, lr 0.01, L2 $5{\cdot}10^{-4}$, 2000 epochs. \\
GPR-GNN~\citep{chien2021adaptiveuniversalgeneralizedpagerank} & Two-layer MLP feature extractor (width 512), then learned GPR propagation: $K=10$, $\alpha=0.1$, personalised PageRank initialisation; feature dropout 0.5, propagation dropout 0.5. $z_v = \sum_{k=0}^{K}\gamma_k H^{(k)}$ (512-dim). & Adam, lr 0.002, wd $5{\cdot}10^{-4}$ on the linear layers and 0 on $\gamma$, 1000 epochs. \\
ACM-GCN~\citep{luan2022revisitingheterophilygnns} & Plain \texttt{acmgcn}: one layer, one hop; three channels (low-pass aggregation, high-pass diversification, identity) mixed per node by learned attention; \texttt{variant=0}, \texttt{structure\_info=0}; width 512. $z_v =$ the mixed representation (512-dim). & AdamW, lr 0.01, wd $10^{-3}$, dropout 0.5, 500 epochs (the authors' large-scale protocol). \\
LINKX~\citep{lim2021largescalelearningnonhomophilous} & PyTorch Geometric \texttt{LINKX}: separate adjacency-row and feature branches combined by a 2-layer MLP; width 512; one edge layer, one node layer, one initial layer per branch; inner activation and inner dropout off; dropout 0.5. $z_v =$ the combined representation (512-dim). & AdamW, lr 0.01, wd $10^{-3}$, 500 epochs. \\
\bottomrule
\end{tabular}
\caption{Fixed encoder and ERM training settings. Here, lr is learning rate
and wd is weight decay. The shared classification
head and the recovery-preparation settings are given in
Sections~\ref{sec:preps} and~\ref{sec:model-training}.}
\label{tab:models}
\end{table}

\noindent The Platonov backbone and its optimiser settings follow the
implementation of \citet{platonov2024criticallookevaluationgnns}.\footnote{\url{https://github.com/yandex-research/heterophilous-graphs}}
The model-specific architectures and baseline settings follow the H2GCN,
GPR-GNN, ACM-GCN, and LINKX source releases.\footnote{\url{https://github.com/GemsLab/H2GCN}}\footnote{\url{https://github.com/jianhao2016/GPRGNN}}\footnote{\url{https://github.com/SitaoLuan/ACM-GNN}}\footnote{\url{https://github.com/CUAI/Non-Homophily-Large-Scale}}
LINKX uses the PyTorch Geometric implementation
\citep{fey2019fastgraphrepresentationlearning}. The shared head, uniform
representation widths, and nested recovery preparations are study-specific
choices described in the main text.

All models receive the same simple undirected graph. We symmetrise directed edges and collapse duplicate edges before applying each model's self-loop convention. We calculate local homophily on the resulting undirected graph, following the convention in \citet[Appendix~N]{mao2023demystifyingstructuraldisparitygraph}.

The GroupDRO implementation uses full-batch group losses. If \(L_g\) is the
mean training cross-entropy for group \(g\), the group weights start uniformly
and are updated after each step as
\[
q_g\leftarrow
\frac{q_g\exp(0.01L_g)}{\sum_j q_j\exp(0.01L_j)}.
\]
The optimised loss is \(\sum_g q_gL_g\). We use no generalisation adjustment.

\section{Training Hardware and Computational Cost}
\label{app:training-hardware}

Training used NVIDIA A100 (80 GB) and NVIDIA GeForce RTX 3080 (16 GB) hardware. All 4,200 main-panel runs used CUDA and one GPU per run.

\begin{table}[H]
  \centering
  \caption{Total GPU-hours by training procedure.}
  \label{tab:training-gpu-hours}
  \begin{tabular}{@{}lrr@{}}
    \toprule
    Procedure & Runs & GPU-hours \\
    \midrule
    \(\Aerm\)  & 1,050 & 179.714 \\
    \(\Alast\) & 1,050 & 19.095 \\
    \(\Ahead\) & 1,050 & 53.941 \\
    \(\Afull\) & 1,050 & 177.189 \\
    \midrule
    Total      & 4,200 & 429.939 \\
    \bottomrule
  \end{tabular}
\end{table}

GPU-hours are calculated from the recorded optimisation-loop duration for
each run. The measurement excludes data loading, preprocessing,
the one-off frozen-feature computation for \(\Alast\) and \(\Ahead\), final
test evaluation, and serialisation.

\section{Local-Homophily Bins}\label{app:local-homophily-bins}

We start by assigning nodes to four base bins: \([0,0.25)\), \([0.25,0.5)\), \([0.5,0.75)\), and \([0.75,1]\). A bin containing fewer than 1\% of nodes is merged into an adjacent bin. Specifically, we repeatedly select the smallest under-threshold bin, with ties resolved towards lower \(h_v\). We then select the larger adjacent bin to merge with. Remaining ties are resolved first towards the dataset mean and then towards lower \(h_v\).

Across the five datasets, the merge rule triggers only once: on Roman-empire, the 81 nodes in \([0.75,1]\) merge with \([0.5,0.75)\).

\begin{table}[H]
    \centering
    \begin{tabular}{@{}llrr}
        \toprule
        Dataset & Bin & Nodes & \% of nodes\\
        \midrule
        PubMed & $\mathbf{[0,0.25)}$ & 2{,}696 & 13.7\\
        (homophilic, mean $h_v$ 0.792) & $[0.25,0.5)$ & 524 & 2.7\\
         & $[0.5,0.75)$ & 1{,}891 & 9.6\\
         & $[0.75,1]$ & 14{,}606 & 74.1\\
        \midrule
        \texttt{ogbn-arxiv} & $\mathbf{[0,0.25)}$ & 30{,}364 & 17.9\\
        (homophilic, mean $h_v$ 0.635) & $[0.25,0.5)$ & 20{,}524 & 12.1\\
         & $[0.5,0.75)$ & 35{,}690 & 21.1\\
         & $[0.75,1]$ & 82{,}765 & 48.9\\
        \midrule
        Roman-empire & $[0,0.25)$ & 20{,}297 & 89.6\\
        (heterophilic, mean $h_v$ 0.046) & $[0.25,0.5)$ & 1{,}353 & 6.0\\
         & $\mathbf{[0.5,1]}$ (merged) & 1{,}012 & 4.5\\
        \midrule
        Amazon-ratings & $[0,0.25)$ & 9{,}465 & 38.6\\
        (heterophilic, mean $h_v$ 0.376) & $[0.25,0.5)$ & 6{,}647 & 27.1\\
         & $[0.5,0.75)$ & 5{,}437 & 22.2\\
         & $\mathbf{[0.75,1]}$ & 2{,}943 & 12.0\\
        \midrule
        \texttt{arxiv-year} & $[0,0.25)$ & 95{,}006 & 56.1\\
        (heterophilic, mean $h_v$ 0.289) & $[0.25,0.5)$ & 33{,}206 & 19.6\\
         & $[0.5,0.75)$ & 19{,}262 & 11.4\\
         & $\mathbf{[0.75,1]}$ & 21{,}869 & 12.9\\
        \bottomrule
    \end{tabular}
    \caption{
        The local homophily bins used throughout. \textbf{Bold} type marks the rare \(h_v\) tail---the low \(h_v\) bin on homophilic datasets and the high-\(h_v\) bin on heterophilic datasets.
    }
    \label{tab:groups}
\end{table}

\section{RQ1 Appendix}
\label{app:rq1}

\begin{figure}[H]
  \centering
  \includegraphics[width=\linewidth]{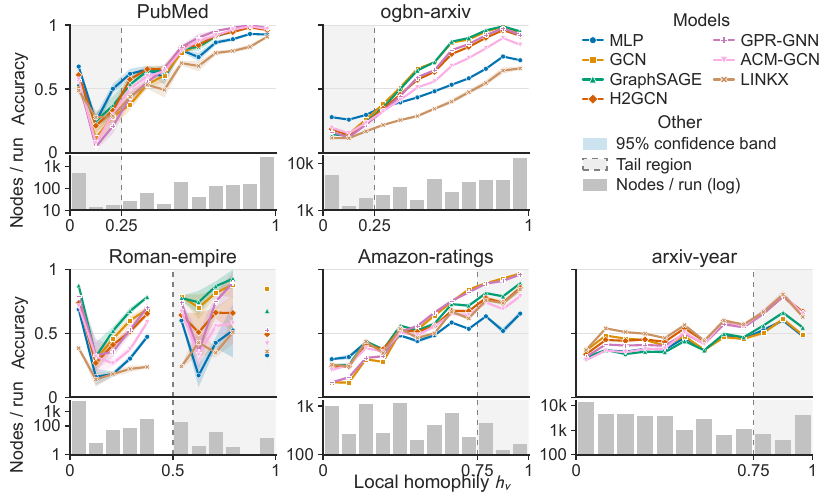}
  \caption{
    \textbf{RQ1}. Average test accuracy across 12 equal-width local-homophily bins. Grey bars show the mean test-node count per run for each bin, and grey regions mark each dataset's rare homophily tail. Roman-empire accuracy is omitted where a bin lacks test nodes in at least one split.
    }
  \label{fig:rq1-continuous-accuracy}
\end{figure}

\subsection{Macro-F1 Tail Gap}

\begin{table}[H]
  \centering
  \caption{ERM signed tail gap: whole-test macro-F1 minus tail macro-F1.
  Each cell gives the mean paired difference over 30 matched split--seed runs
  and its 95\% percentile-bootstrap interval. Positive values indicate lower
  tail performance; grey cells contain zero.}
  \label{tab:erm-tail-gap}
  \begin{tabular}{@{}lccc@{}}
    \toprule
    & & \multicolumn{2}{c}{Low-homophily tail} \\
    \cmidrule(lr){3-4}
    Model & & PubMed & \texttt{ogbn-arxiv} \\
    \midrule
    MLP             & & \gaincell{+.208}{[.203,.213]} & \gaincell{+.102}{[.102,.103]} \\
    GCN             & & \gaincell{+.364}{[.358,.369]} & \gaincell{+.332}{[.331,.333]} \\
    GraphSAGE-mean  & & \gaincell{+.353}{[.348,.358]} & \gaincell{+.356}{[.354,.357]} \\
    H2GCN-2         & & \gaincell{+.299}{[.292,.305]} & \gaincell{+.301}{[.300,.303]} \\
    GPR-GNN         & & \gaincell{+.350}{[.345,.356]} & \gaincell{+.305}{[.303,.306]} \\
    ACM-GCN         & & \gaincell{+.332}{[.327,.338]} & \gaincell{+.225}{[.223,.226]} \\
    LINKX           & & \gaincell{+.332}{[.313,.351]} & \gaincell{+.173}{[.169,.177]} \\
    \bottomrule
    \noalign{\vskip 3pt}
    \toprule
    & \multicolumn{3}{c}{High-homophily tail} \\
    \cmidrule(lr){2-4}
    Model & Roman-empire & Amazon-ratings & \texttt{arxiv-year} \\
    \midrule
    MLP             & \uncertaingaincell{+.022}{[-.001,.044]} & \gaincell{-.044}{[-.052,-.037]} & \gaincell{+.028}{[.026,.030]} \\
    GCN             & \gaincell{-.025}{[-.045,-.006]} & \gaincell{-.408}{[-.425,-.390]} & \gaincell{+.070}{[.067,.072]} \\
    GraphSAGE-mean  & \gaincell{+.160}{[.134,.188]} & \gaincell{-.196}{[-.208,-.185]} & \uncertaingaincell{.000}{[-.002,.001]} \\
    H2GCN-2         & \gaincell{+.097}{[.075,.120]} & \gaincell{-.191}{[-.208,-.176]} & \uncertaingaincell{.000}{[-.003,.003]} \\
    GPR-GNN         & \gaincell{+.077}{[.056,.097]} & \gaincell{-.419}{[-.434,-.403]} & \gaincell{-.015}{[-.018,-.012]} \\
    ACM-GCN         & \gaincell{+.034}{[.016,.053]} & \gaincell{-.208}{[-.225,-.190]} & \gaincell{-.024}{[-.027,-.022]} \\
    LINKX           & \gaincell{+.046}{[.035,.056]} & \gaincell{-.157}{[-.175,-.141]} & \gaincell{+.069}{[.065,.073]} \\
    \bottomrule
  \end{tabular}
\end{table}

\subsection{Whole-Test and Tail Accuracy}

\begin{table}[H]
  \centering
  \caption{ERM accuracy gap, whole-test accuracy minus tail
  accuracy. Each cell gives the mean paired difference over 30 matched
  split--seed runs and its 95\% percentile-bootstrap interval. Positive
  values indicate lower tail accuracy; grey cells contain zero.}
  \label{tab:erm-tail-gap-accuracy}
  \begin{tabular}{@{}lccc@{}}
    \toprule
    & & \multicolumn{2}{c}{Low-homophily tail} \\
    \cmidrule(lr){3-4}
    Model & & PubMed & \texttt{ogbn-arxiv} \\
    \midrule
    MLP              & & \gaincell{+.211}{[.206,.217]} & \gaincell{+.270}{[.269,.271]} \\
    GCN              & & \gaincell{+.369}{[.363,.375]} & \gaincell{+.520}{[.519,.521]} \\
    GraphSAGE-mean   & & \gaincell{+.359}{[.354,.365]} & \gaincell{+.555}{[.553,.557]} \\
    H2GCN-2          & & \gaincell{+.302}{[.295,.308]} & \gaincell{+.493}{[.492,.495]} \\
    GPR-GNN          & & \gaincell{+.355}{[.349,.360]} & \gaincell{+.517}{[.515,.519]} \\
    ACM-GCN          & & \gaincell{+.338}{[.332,.345]} & \gaincell{+.428}{[.427,.430]} \\
    LINKX            & & \gaincell{+.337}{[.317,.357]} & \gaincell{+.310}{[.298,.322]} \\
    \bottomrule
    \noalign{\vskip 3pt}
    \toprule
    & \multicolumn{3}{c}{High-homophily tail} \\
    \cmidrule(lr){2-4}
    Model & Roman-empire & Amazon-ratings & \texttt{arxiv-year} \\
    \midrule
    MLP              & \gaincell{+.111}{[.099,.121]} & \gaincell{-.159}{[-.165,-.153]} & \gaincell{-.124}{[-.127,-.122]} \\
    GCN              & \gaincell{-.061}{[-.069,-.053]} & \gaincell{-.438}{[-.442,-.434]} & \gaincell{-.075}{[-.085,-.065]} \\
    GraphSAGE-mean   & \gaincell{+.077}{[.069,.086]} & \gaincell{-.304}{[-.310,-.299]} & \gaincell{-.181}{[-.188,-.174]} \\
    H2GCN-2          & \gaincell{+.089}{[.079,.099]} & \gaincell{-.292}{[-.298,-.285]} & \gaincell{-.232}{[-.236,-.229]} \\
    GPR-GNN          & \gaincell{+.052}{[.044,.059]} & \gaincell{-.424}{[-.429,-.420]} & \gaincell{-.249}{[-.257,-.240]} \\
    ACM-GCN          & \gaincell{+.065}{[.058,.072]} & \gaincell{-.273}{[-.280,-.265]} & \gaincell{-.273}{[-.278,-.268]} \\
    LINKX            & \gaincell{+.098}{[.086,.110]} & \gaincell{-.282}{[-.288,-.275]} & \gaincell{-.170}{[-.178,-.162]} \\
    \bottomrule
  \end{tabular}
\end{table}

\section{RQ1a: Within-Class Analyses and Class-Composition Adjustment}
\label{app:rq2-within-class}

\subsection{Within-class accuracy.}
Within each run, we calculate tail and non-tail accuracy separately for every true class represented in both subsets. We subtract tail accuracy from non-tail accuracy within each class, then average the class-wise differences equally.

\begin{table}[H]
    \centering
    \caption{
        Per-class non-tail accuracy minus tail accuracy, averaged equally over classes. Each cell reports the mean over the 30 runs and the 95\% percentile-bootstrap interval. Grey cells denote confidence intervals that contain zero.
    }
    \label{tab:rq2-class-macro-accuracy}
    \begin{tabular}{@{}lccc@{}}
      \toprule
      \multicolumn{2}{@{}l}{} & \multicolumn{2}{c}{Low-homophily tail} \\
      \cmidrule(lr){3-4}
      \multicolumn{2}{@{}l}{Model} & PubMed & \texttt{ogbn-arxiv} \\
      \midrule
      \multicolumn{2}{@{}l}{MLP}            & \gaincell{+.229}{[.224,.235]} & \gaincell{+.151}{[.149,.152]} \\
      \multicolumn{2}{@{}l}{GCN}            & \gaincell{+.419}{[.412,.428]} & \gaincell{+.498}{[.496,.501]} \\
      \multicolumn{2}{@{}l}{GraphSAGE-mean} & \gaincell{+.406}{[.399,.414]} & \gaincell{+.516}{[.513,.518]} \\
      \multicolumn{2}{@{}l}{H2GCN-2}        & \gaincell{+.344}{[.336,.352]} & \gaincell{+.415}{[.412,.417]} \\
      \multicolumn{2}{@{}l}{GPR-GNN}        & \gaincell{+.401}{[.393,.408]} & \gaincell{+.401}{[.398,.404]} \\
      \multicolumn{2}{@{}l}{ACM-GCN}        & \gaincell{+.379}{[.371,.387]} & \gaincell{+.302}{[.300,.304]} \\
      \multicolumn{2}{@{}l}{LINKX}          & \gaincell{+.391}{[.367,.414]} & \gaincell{+.293}{[.286,.300]} \\
      \bottomrule
      \noalign{\vskip 3pt}
      \toprule
      & \multicolumn{3}{c}{High-homophily tail} \\
      \cmidrule(lr){2-4}
      Model & Roman-empire & Amazon-ratings & \texttt{arxiv-year} \\
      \midrule
      MLP              & \gaincell{-.050}{[-.066,-.035]} & \gaincell{-.210}{[-.229,-.192]} & \gaincell{+.004}{[.003,.006]} \\
      GCN              & \gaincell{-.153}{[-.174,-.130]} & \gaincell{-.566}{[-.573,-.559]} & \gaincell{+.063}{[.060,.066]} \\
      GraphSAGE-mean   & \gaincell{+.071}{[.048,.092]} & \gaincell{-.376}{[-.387,-.363]} & \gaincell{-.018}{[-.020,-.015]} \\
      H2GCN-2          & \uncertaingaincell{-.010}{[-.029,.008]} & \gaincell{-.316}{[-.334,-.299]} & \gaincell{+.016}{[.012,.019]} \\
      GPR-GNN          & \gaincell{-.015}{[-.030,-.001]} & \gaincell{-.529}{[-.539,-.519]} & \gaincell{-.005}{[-.007,-.003]} \\
      ACM-GCN          & \gaincell{-.049}{[-.062,-.035]} & \gaincell{-.310}{[-.332,-.289]} & \gaincell{-.004}{[-.007,-.001]} \\
      LINKX            & \gaincell{-.052}{[-.067,-.039]} & \gaincell{-.328}{[-.354,-.303]} & \gaincell{+.025}{[.021,.030]} \\
      \bottomrule
    \end{tabular}
\end{table}

\subsection{Class-adjusted RQ1 gaps.}
A tail gap may arise because the whole test set and the tail
contain different proportions of easier and harder classes. We therefore employ the following adjustment procedure for class composition, following \citep{keiding2014standardization}.

For a run \(r\), $C_r$ denotes the set of all classes present in the tail. We calculate the whole-test score using only nodes whose true class is in $C_r$. Once restricted to these classes, the whole test set by definition already has the reference class composition and requires no further reweighting.

Following the confusion-matrix construction of \citet[Section~3.4]{opitz2024closer}, for $S \in \{ \mathrm{whole},\mathrm{tail} \}$ let $M^S_{r,ij}$ be the number of nodes in subset $S$ with true class $i \in C_r$ that are predicted as class $j$, where $j$ ranges over every class in the dataset. Let $n^S_{r,i}$ be the number of nodes of class $i$ in subset $S$, and let $w_{r,i}$ be the proportion of class $i$ among the selected whole-test nodes:

\[
    n^S_{r,i} = \sum_j M^S_{r,ij},
    \qquad
    w_{r,i}
    =
    \frac{
        n^{\mathrm{whole}}_{r,i}
    }{
        \sum_{\ell \in C_r}n^{\mathrm{whole}}_{r, \ell}
    }.
\]

We construct the adjusted tail confusion matrix as
\[
    \widetilde M^{\mathrm{tail}}_{r,ij}
    =
    w_{r,i}
    \frac{
        M^{\mathrm{tail}}_{r,ij}
    }{
        n^{\mathrm{tail}}_{r,i}
    }.
\]

Dividing by $n^{\mathrm{tail}}_{r,i}$ retains the tail's distribution of
predictions within class $i$. Multiplying by $w_{r,i}$ gives that class its
whole-test proportion. \textbf{The adjusted tail matrix therefore has the class
composition of the whole test set while retaining the prediction patterns
observed in the tail}.

A prediction outside $C_r$ therefore still counts as an error, although nodes
whose true class lies outside $C_r$ are excluded from both evaluations.

We calculate accuracy and macro-F1 from the adjusted tail matrix and compare
them with the corresponding whole-test scores. Accuracy is the total weight
assigned to correct predictions. Macro-F1 is the unweighted mean of the
one-vs-rest F1 scores for classes in $C_r$. For either metric, the adjusted gap \(G^{\mathrm{adj}}_r\)
in run $r$ is
\[
    G^{\mathrm{adj}}_r
    =
    \operatorname{Metric}_{\mathrm{whole},r}
    -
    \operatorname{Metric}^{\mathrm{adj}}_{\mathrm{tail},r}.
\]

If every whole-test class appears in the tail, the whole-test score is the
original unadjusted RQ1 whole-test score. A whole-test class absent from the tail is
excluded from both evaluations. The adjusted gap can therefore differ from the
original RQ1 gap through tail reweighting and, where class support differs,
through the restricted class set. Note that \emph{adjusted} here refers only to this
statistical standardisation and carries no causal interpretation.

\begin{table}[H]
  \centering
  \caption{
    Class-adjusted accuracy tail gaps for all models on all datasets. Entries are the mean and 95\% bootstrap interval across 30 runs. Grey cells include zero, and $\dagger$ marks cells whose unadjusted accuracy tail gap includes zero.
  }
  \label{tab:rq2-adjusted-accuracy}
  \begin{tabular}{@{}lccc@{}}
    \toprule
    \multicolumn{2}{@{}l}{} & \multicolumn{2}{c}{Low-homophily tail} \\
    \cmidrule(lr){3-4}
    \multicolumn{2}{@{}l}{Model} & PubMed & \texttt{ogbn-arxiv} \\
    \midrule
    \multicolumn{2}{@{}l}{MLP}            & \gaincell{+.197}{[.191,.203]} & \gaincell{+.188}{[.187,.190]} \\
    \multicolumn{2}{@{}l}{GCN}            & \gaincell{+.376}{[.369,.383]} & \gaincell{+.498}{[.497,.500]} \\
    \multicolumn{2}{@{}l}{GraphSAGE-mean} & \gaincell{+.366}{[.360,.373]} & \gaincell{+.524}{[.523,.527]} \\
    \multicolumn{2}{@{}l}{H2GCN-2}        & \gaincell{+.306}{[.298,.313]} & \gaincell{+.427}{[.426,.429]} \\
    \multicolumn{2}{@{}l}{GPR-GNN}        & \gaincell{+.354}{[.346,.361]} & \gaincell{+.451}{[.449,.452]} \\
    \multicolumn{2}{@{}l}{ACM-GCN}        & \gaincell{+.335}{[.329,.342]} & \gaincell{+.353}{[.351,.354]} \\
    \multicolumn{2}{@{}l}{LINKX}          & \gaincell{+.345}{[.325,.365]} & \gaincell{+.282}{[.274,.292]} \\
    \bottomrule
    \noalign{\vskip 3pt}
    \toprule
    & \multicolumn{3}{c}{High-homophily tail} \\
    \cmidrule(lr){2-4}
    Model & Roman-empire & Amazon-ratings & \texttt{arxiv-year} \\
    \midrule
    MLP              & \uncertaingaincell{+.008}{[-.013,.029]} & \gaincell{-.137}{[-.145,-.128]} & \gaincell{+.017}{[.014,.018]} \\
    GCN              & \gaincell{-.044}{[-.071,-.020]} & \gaincell{-.437}{[-.441,-.433]} & \gaincell{+.088}{[.085,.091]} \\
    GraphSAGE-mean   & \gaincell{+.115}{[.086,.146]} & \gaincell{-.284}{[-.292,-.277]} & \gaincell{+.004}{[.001,.006]} \\
    H2GCN-2          & \gaincell{+.058}{[.031,.084]} & \gaincell{-.254}{[-.261,-.247]} & \gaincell{+.052}{[.048,.055]} \\
    GPR-GNN          & \gaincell{+.031}{[.009,.052]} & \gaincell{-.401}{[-.406,-.395]} & \gaincell{+.023}{[.021,.025]} \\
    ACM-GCN          & \uncertaingaincell{-.005}{[-.023,.015]} & \gaincell{-.238}{[-.246,-.229]} & \gaincell{+.021}{[.018,.024]} \\
    LINKX            & \uncertaingaincell{-.021}{[-.044,.001]} & \gaincell{-.265}{[-.274,-.255]} & \gaincell{+.057}{[.052,.061]} \\
    \bottomrule
  \end{tabular}
\end{table}

\begin{table}[H]
  \centering
  \caption{
    Class-adjusted macro-F1 tail gaps for all models on all datasets. Entries are the mean and 95\% bootstrap interval across 30 runs. Grey cells include zero, and $\dagger$ marks cells whose unadjusted macro-F1 tail gap includes zero.
  }
  \label{tab:rq2-adjusted-macro-f1}
  \begin{tabular}{@{}lccc@{}}
    \toprule
    \multicolumn{2}{@{}l}{} & \multicolumn{2}{c}{Low-homophily tail} \\
    \cmidrule(lr){3-4}
    \multicolumn{2}{@{}l}{Model} & PubMed & \texttt{ogbn-arxiv} \\
    \midrule
    \multicolumn{2}{@{}l}{MLP}            & \gaincell{+.197}{[.192,.203]} & \gaincell{+.106}{[.105,.107]} \\
    \multicolumn{2}{@{}l}{GCN}            & \gaincell{+.364}{[.358,.371]} & \gaincell{+.350}{[.348,.351]} \\
    \multicolumn{2}{@{}l}{GraphSAGE-mean} & \gaincell{+.350}{[.345,.357]} & \gaincell{+.366}{[.364,.367]} \\
    \multicolumn{2}{@{}l}{H2GCN-2}        & \gaincell{+.298}{[.291,.305]} & \gaincell{+.299}{[.298,.301]} \\
    \multicolumn{2}{@{}l}{GPR-GNN}        & \gaincell{+.348}{[.340,.354]} & \gaincell{+.303}{[.301,.304]} \\
    \multicolumn{2}{@{}l}{ACM-GCN}        & \gaincell{+.327}{[.321,.334]} & \gaincell{+.222}{[.221,.223]} \\
    \multicolumn{2}{@{}l}{LINKX}          & \gaincell{+.335}{[.316,.355]} & \gaincell{+.179}{[.175,.183]} \\
    \bottomrule
    \noalign{\vskip 3pt}
    \toprule
    & \multicolumn{3}{c}{High-homophily tail} \\
    \cmidrule(lr){2-4}
    Model & Roman-empire & Amazon-ratings & \texttt{arxiv-year} \\
    \midrule
    MLP              & \gaincell{-.022^{\dagger}}{[-.036,-.009]} & \gaincell{-.185}{[-.201,-.170]} & \gaincell{+.015}{[.013,.016]} \\
    GCN              & \gaincell{-.096}{[-.121,-.073]} & \gaincell{-.509}{[-.515,-.503]} & \gaincell{+.061}{[.059,.062]} \\
    GraphSAGE-mean   & \gaincell{+.106}{[.081,.134]} & \gaincell{-.324}{[-.336,-.312]} & \uncertaingaincell{+.001^{\dagger}}{[.000,.003]} \\
    H2GCN-2          & \gaincell{+.030}{[.010,.049]} & \gaincell{-.285}{[-.302,-.269]} & \gaincell{+.025^{\dagger}}{[.021,.028]} \\
    GPR-GNN          & \gaincell{+.030}{[.013,.045]} & \gaincell{-.484}{[-.492,-.476]} & \gaincell{+.013}{[.011,.014]} \\
    ACM-GCN          & \gaincell{-.020}{[-.033,-.008]} & \gaincell{-.291}{[-.312,-.269]} & \gaincell{+.009}{[.007,.011]} \\
    LINKX            & \gaincell{-.038}{[-.052,-.023]} & \gaincell{-.289}{[-.309,-.267]} & \gaincell{+.058}{[.052,.064]} \\
    \bottomrule
  \end{tabular}
\end{table}

\section{RQ1b Tail Gaps of Aggregation models vs MLP}
\label{app:rq3-aggregation-vs-mlp}

\begin{figure}[H]
  \centering
  \includegraphics[width=\linewidth]
  {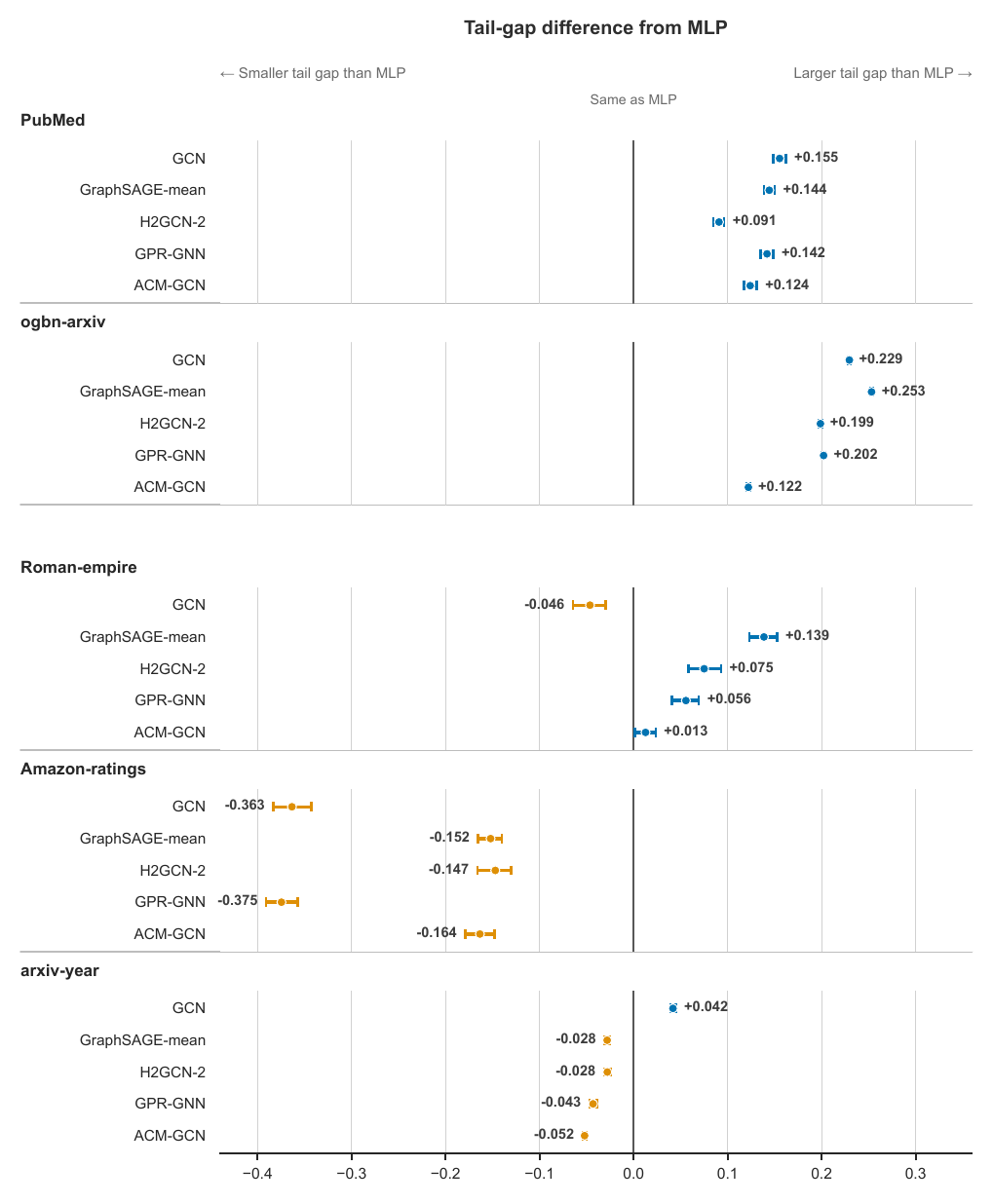}
  \caption{
    Macro-F1 tail gap of the five aggregation GNNs minus the tail gap of the graphless MLP model for all five datasets. Each point is the mean over 30 matched split--seed differences; error bars show 95\% percentile-bootstrap intervals.
    }
  \label{fig:rq3-tail-gap-contrast}
\end{figure}

\begin{figure}[H]
  \centering
  \includegraphics[width=\linewidth]{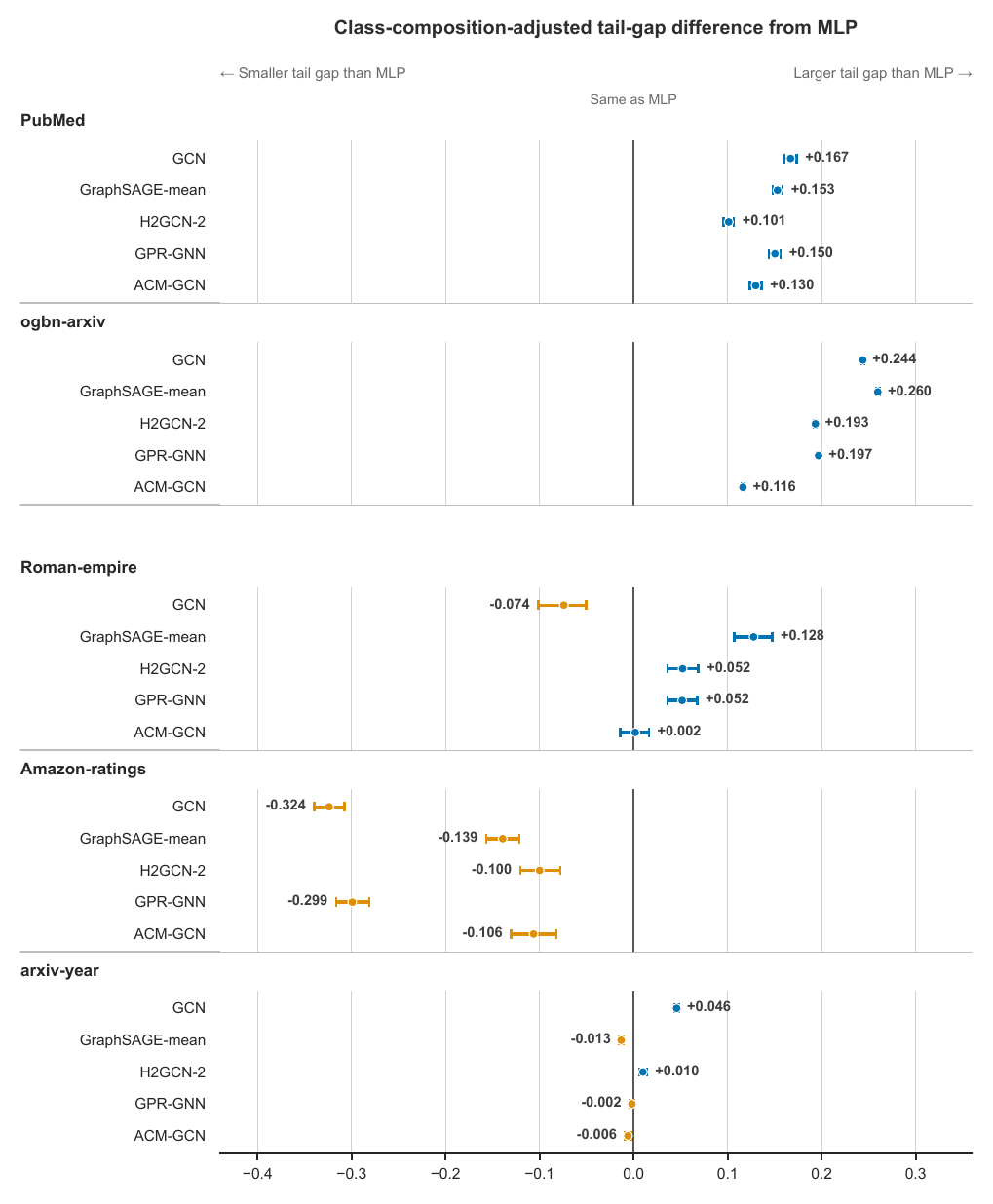}
  \caption{
    Class-composition adjusted macro-F1 tail gap of the five aggregation GNNs minus the tail gap of the graphless MLP model for all five datasets. Each point is the mean over 30 matched split--seed differences; error bars show 95\% percentile-bootstrap intervals.}
  \label{fig:rq3-aggregation-vs-mlp-class-adjusted}
\end{figure}

Here we present more details on the differences in tail gaps between aggregation-based GNNs and the baseline MLP model. Figure~\ref{fig:rq3-tail-gap-contrast} plots this difference in macro-F1 tail gaps for all aggregation models on all datasets. A larger tail gap indicates a larger difference in model performance on the rare homophily tail compared to the test set as a whole. Positive tail gaps indicate that the model scored worse on the tail. Figure~\ref{fig:rq3-aggregation-vs-mlp-class-adjusted} shows the same results adjusted for class composition (cf. Appendix~\ref{app:rq2-within-class}).

\section{RQ1c: Heterophily-Designed Versus Standard Aggregation Models}
\label{app:rq4-designed-vs-standard}

\begin{table}[H]
  \centering
  \caption{
    \textbf{RQ1c}. Macro-F1 tail-gap comparison between each heterophily-designed model and each standard aggregation baseline. Each cell is heterophily-model gap minus standard-baseline gap, with its 95\% percentile-bootstrap interval over the 30 matched split--seed runs. Negative values indicate a smaller tail gap for the heterophily-designed model; grey cells contain zero and are not statistically significant.
    }
  \label{tab:rq4-designed-vs-standard}
  \setlength{\tabcolsep}{5pt}
  \begin{tabular}{@{}llccc@{}}
    \toprule
    & & & \multicolumn{2}{c}{Low-homophily tail} \\
    \cmidrule(lr){4-5}
    Designed model & Standard baseline & & PubMed & \texttt{ogbn-arxiv} \\
    \midrule
    \multirow{2}{*}{H2GCN-2} & GCN & & \gaincell{-.064}{[-.071,-.059]} & \gaincell{-.031}{[-.033,-.029]} \\
     & GraphSAGE-mean & & \gaincell{-.053}{[-.059,-.048]} & \gaincell{-.054}{[-.056,-.053]} \\
    \addlinespace[2pt]
    \multirow{2}{*}{GPR-GNN} & GCN & & \gaincell{-.013}{[-.019,-.008]} & \gaincell{-.027}{[-.029,-.025]} \\
     & GraphSAGE-mean & & \uncertaingaincell{-.002}{[-.009,+.004]} & \gaincell{-.051}{[-.053,-.049]} \\
    \addlinespace[2pt]
    \multirow{2}{*}{ACM-GCN} & GCN & & \gaincell{-.031}{[-.039,-.023]} & \gaincell{-.107}{[-.109,-.105]} \\
     & GraphSAGE-mean & & \gaincell{-.020}{[-.025,-.015]} & \gaincell{-.131}{[-.133,-.129]} \\
    \bottomrule
    \noalign{\vskip 3pt}
    \toprule
    & & \multicolumn{3}{c}{High-homophily tail} \\
    \cmidrule(lr){3-5}
    Designed model & Standard baseline & Roman-empire & Amazon-ratings & \texttt{arxiv-year} \\
    \midrule
    \multirow{2}{*}{H2GCN-2} & GCN & \gaincell{+.121}{[+.104,+.140]} & \gaincell{+.216}{[+.189,+.243]} & \gaincell{-.070}{[-.074,-.065]} \\
     & GraphSAGE-mean & \gaincell{-.064}{[-.077,-.051]} & \uncertaingaincell{+.005}{[-.012,+.023]} & \uncertaingaincell{.000}{[-.003,+.004]} \\
    \addlinespace[2pt]
    \multirow{2}{*}{GPR-GNN} & GCN & \gaincell{+.102}{[+.090,+.114]} & \uncertaingaincell{-.011}{[-.030,+.009]} & \gaincell{-.085}{[-.089,-.080]} \\
     & GraphSAGE-mean & \gaincell{-.083}{[-.097,-.070]} & \gaincell{-.222}{[-.240,-.203]} & \gaincell{-.015}{[-.019,-.011]} \\
    \addlinespace[2pt]
    \multirow{2}{*}{ACM-GCN} & GCN & \gaincell{+.059}{[+.047,+.073]} & \gaincell{+.200}{[+.169,+.231]} & \gaincell{-.094}{[-.097,-.091]} \\
     & GraphSAGE-mean & \gaincell{-.126}{[-.139,-.113]} & \uncertaingaincell{-.011}{[-.033,+.009]} & \gaincell{-.024}{[-.027,-.021]} \\
    \bottomrule
  \end{tabular}
\end{table}

\begin{figure}[H]
  \centering
  \includegraphics[width=\linewidth]{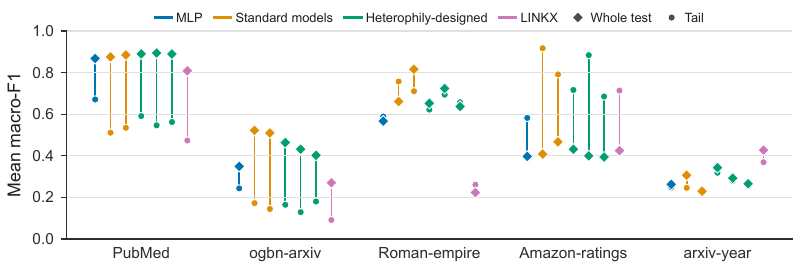}
  \caption{\textbf{RQ1c}. Class-composition-standardised whole-test and tail
  macro-F1 for all models on all datasets, with models coloured by model
  family. The tail is reweighted to the whole-test proportions of true classes
  represented in the tail; the whole-test score uses the same class set.
  Within each dataset, models are ordered from left to right as MLP, GCN,
  GraphSAGE-mean, H2GCN-2, GPR-GNN, ACM-GCN, and LINKX. Blue denotes MLP,
  orange the standard aggregation models, green the heterophily-designed
  aggregation models, and purple LINKX.}
  \label{fig:rq4-class-adjusted-whole-tail-scores}
\end{figure}

\paragraph{Class-composition-standardised analysis.}
Figure~\ref{fig:rq4-class-adjusted-whole-tail-scores} repeats the
absolute-score comparison after the class-composition standardisation defined
in Appendix~\ref{app:rq2-within-class}.

\section{RQ2: Tail and Non-Tail Performance Across Retraining Procedures}
\label{app:rq5-preparation-scores}

\begin{table}[H]
  \centering
  \caption{
    \textbf{RQ2a}. Change in tail macro-F1 after last layer retraining, \(\Alast-\Aerm\). Each cell gives the mean difference over the 30 runs and its 95\% percentile-bootstrap interval. Grey cells contain zero and hence are not statistically significant.}
  \label{tab:last-tail-gain}
  \begin{tabular}{@{}lccc@{}}
      \toprule
      & & \multicolumn{2}{c}{Low-homophily tail} \\
      \cmidrule(lr){3-4}
      Model & & PubMed & \texttt{ogbn-arxiv} \\
      \midrule
      MLP             & & \gaincell{+.016}{[.011,.020]} & \gaincell{+.008}{[.006,.009]} \\
      GCN             & & \gaincell{+.025}{[.016,.034]} & \gaincell{+.036}{[.035,.038]} \\
      GraphSAGE-mean  & & \gaincell{+.116}{[.105,.127]} & \gaincell{+.053}{[.051,.055]} \\
      H2GCN-2         & & \gaincell{+.057}{[.042,.072]} & \gaincell{+.076}{[.074,.078]} \\
      GPR-GNN         & & \gaincell{+.022}{[.014,.030]} & \gaincell{+.093}{[.091,.096]} \\
      ACM-GCN         & & \gaincell{+.005}{[.002,.009]} & \gaincell{+.072}{[.069,.074]} \\
      LINKX           & & \gaincell{+.028}{[.014,.044]} & \gaincell{+.018}{[.015,.020]} \\
      \bottomrule
    \noalign{\vskip 3pt}
      \toprule
      & \multicolumn{3}{c}{High-homophily tail} \\
      \cmidrule(lr){2-4}
      Model & Roman-empire & Amazon-ratings & \texttt{arxiv-year} \\
      \midrule
      MLP             & \gaincell{+.068}{[.060,.077]} & \uncertaingaincell{-.005}{[-.018,.007]} & \gaincell{+.016}{[.012,.020]} \\
      GCN             & \uncertaingaincell{.000}{[-.015,.013]} & \gaincell{-.245}{[-.275,-.215]} & \gaincell{+.041}{[.036,.048]} \\
      GraphSAGE-mean  & \uncertaingaincell{.000}{[-.006,.005]} & \gaincell{-.017}{[-.026,-.009]} & \gaincell{+.047}{[.044,.050]} \\
      H2GCN-2         & \gaincell{+.041}{[.028,.052]} & \gaincell{+.026}{[.001,.049]} & \gaincell{+.007}{[.005,.009]} \\
      GPR-GNN         & \gaincell{+.040}{[.031,.050]} & \uncertaingaincell{-.004}{[-.028,.017]} & \gaincell{+.009}{[.006,.011]} \\
      ACM-GCN         & \gaincell{+.032}{[.018,.045]} & \gaincell{-.020}{[-.036,-.001]} & \gaincell{+.006}{[.004,.008]} \\
      LINKX           & \gaincell{+.050}{[.036,.064]} & \uncertaingaincell{-.001}{[-.014,.012]} & \gaincell{+.016}{[.011,.021]} \\
      \bottomrule
  \end{tabular}
\end{table}

\begin{table}[H]
  \centering
  \caption{
    \textbf{RQ2b}. Change in tail macro-F1 after head retraining, \(\Ahead-\Aerm\). Each cell gives the mean difference over the 30 runs and its 95\% percentile-bootstrap interval. Grey cells contain zero and hence are not statistically significant.}
  \label{tab:head-tail-gain}
  \begin{tabular}{@{}lccc@{}}
      \toprule
      & & \multicolumn{2}{c}{Low-homophily tail} \\
      \cmidrule(lr){3-4}
      Model & & PubMed & \texttt{ogbn-arxiv} \\
      \midrule
      MLP             & & \gaincell{+.021}{[.017,.026]} & \gaincell{+.032}{[.031,.033]} \\
      GCN             & & \gaincell{+.108}{[.099,.118]} & \gaincell{+.090}{[.088,.091]} \\
      GraphSAGE-mean  & & \gaincell{+.156}{[.148,.163]} & \gaincell{+.121}{[.120,.123]} \\
      H2GCN-2         & & \gaincell{+.048}{[.038,.059]} & \gaincell{+.069}{[.067,.072]} \\
      GPR-GNN         & & \gaincell{+.017}{[.009,.024]} & \gaincell{+.129}{[.125,.132]} \\
      ACM-GCN         & & \gaincell{+.018}{[.013,.023]} & \gaincell{+.055}{[.052,.058]} \\
      LINKX           & & \gaincell{+.064}{[.040,.091]} & \gaincell{+.024}{[.021,.027]} \\
      \bottomrule
    \noalign{\vskip 3pt}
      \toprule
      & \multicolumn{3}{c}{High-homophily tail} \\
      \cmidrule(lr){2-4}
      Model & Roman-empire & Amazon-ratings & \texttt{arxiv-year} \\
      \midrule
      MLP             & \gaincell{+.061}{[.051,.070]} & \gaincell{+.018}{[.001,.032]} & \gaincell{+.029}{[.026,.032]} \\
      GCN             & \uncertaingaincell{-.017}{[-.038,.003]} & \uncertaingaincell{+.004}{[-.020,.029]} & \gaincell{+.086}{[.083,.090]} \\
      GraphSAGE-mean  & \uncertaingaincell{-.001}{[-.014,.008]} & \gaincell{+.042}{[.028,.059]} & \gaincell{+.064}{[.062,.067]} \\
      H2GCN-2         & \gaincell{+.045}{[.032,.058]} & \gaincell{+.034}{[.009,.061]} & \gaincell{+.029}{[.024,.034]} \\
      GPR-GNN         & \gaincell{+.054}{[.043,.064]} & \uncertaingaincell{+.002}{[-.014,.019]} & \gaincell{+.010}{[.007,.013]} \\
      ACM-GCN         & \gaincell{+.056}{[.041,.069]} & \uncertaingaincell{-.009}{[-.022,.004]} & \gaincell{+.067}{[.063,.071]} \\
      LINKX           & \gaincell{+.064}{[.043,.084]} & \uncertaingaincell{+.002}{[-.008,.012]} & \gaincell{+.021}{[.015,.026]} \\
      \bottomrule
  \end{tabular}
\end{table}

\begin{table}[H]
  \centering
  \caption{
    \textbf{RQ2c}. Change in tail macro-F1 after full model retraining, \(\Afull-\Aerm\). Each cell gives the mean difference over the 30 runs and its 95\% percentile-bootstrap interval. Grey cells contain zero and hence are not statistically significant.
    }
  \label{tab:full-tail-gain}
  \begin{tabular}{@{}lccc@{}}
      \toprule
      & & \multicolumn{2}{c}{Low-homophily tail} \\
      \cmidrule(lr){3-4}
      Model & & PubMed & \texttt{ogbn-arxiv} \\
      \midrule
      MLP             & & \uncertaingaincell{+.004}{[-.001,.010]} & \gaincell{+.023}{[.022,.025]} \\
      GCN             & & \gaincell{+.122}{[.115,.129]} & \gaincell{+.094}{[.090,.097]} \\
      GraphSAGE-mean  & & \gaincell{+.162}{[.156,.169]} & \gaincell{+.128}{[.126,.130]} \\
      H2GCN-2         & & \gaincell{+.064}{[.056,.072]} & \gaincell{+.081}{[.078,.084]} \\
      GPR-GNN         & & \gaincell{+.095}{[.089,.102]} & \gaincell{+.074}{[.071,.077]} \\
      ACM-GCN         & & \gaincell{+.101}{[.095,.108]} & \gaincell{+.064}{[.061,.066]} \\
      LINKX           & & \gaincell{+.089}{[.070,.108]} & \gaincell{+.031}{[.028,.034]} \\
      \bottomrule
    \noalign{\vskip 3pt}
      \toprule
      & \multicolumn{3}{c}{High-homophily tail} \\
      \cmidrule(lr){2-4}
      Model & Roman-empire & Amazon-ratings & \texttt{arxiv-year} \\
      \midrule
      MLP             & \gaincell{+.057}{[.044,.069]} & \gaincell{-.099}{[-.111,-.087]} & \uncertaingaincell{-.001}{[-.004,.003]} \\
      GCN             & \gaincell{-.039}{[-.062,-.020]} & \gaincell{-.150}{[-.177,-.119]} & \gaincell{+.022}{[.020,.024]} \\
      GraphSAGE-mean  & \gaincell{+.036}{[.027,.046]} & \gaincell{-.084}{[-.105,-.067]} & \gaincell{+.018}{[.016,.019]} \\
      H2GCN-2         & \gaincell{+.084}{[.072,.097]} & \uncertaingaincell{-.011}{[-.039,.017]} & \gaincell{-.017}{[-.022,-.014]} \\
      GPR-GNN         & \gaincell{+.054}{[.043,.065]} & \gaincell{-.287}{[-.327,-.244]} & \gaincell{+.006}{[.003,.009]} \\
      ACM-GCN         & \gaincell{+.042}{[.034,.052]} & \gaincell{-.060}{[-.078,-.042]} & \gaincell{+.015}{[.012,.019]} \\
      LINKX           & \gaincell{+.058}{[.039,.076]} & \gaincell{+.075}{[.050,.101]} & \gaincell{+.061}{[.053,.068]} \\
      \bottomrule
  \end{tabular}
\end{table}

\begin{table}[H]
  \centering
  \caption{\textbf{RQ2a}. Recovered share \(\rho_{\mathrm{last}}\) after final-layer refitting; cells are grey when the 95\% paired-bootstrap interval for the numerator contains zero or \(\hat A_{\mathrm{full}}-\hat A_{\mathrm{ERM}}<0.03\).}
  \label{tab:rho-last-recovered-share}
  \setlength{\tabcolsep}{5.5pt}
  \begin{tabular*}{\linewidth}{@{\extracolsep{\fill}}lccccc@{}}
    \toprule
    & \multicolumn{2}{c}{Low-homophily tail}
    & \multicolumn{3}{c}{High-homophily tail} \\
    \cmidrule(lr){2-3}\cmidrule(lr){4-6}
    Model & PubMed & \texttt{ogbn-arxiv} & Roman-empire & Amazon-ratings & \texttt{arxiv-year} \\
    \midrule
    GCN & $0.208$ & $0.388$ & \cellcolor{black!10}$-0.001$ & \cellcolor{black!10}$1.632$ & \cellcolor{black!10}$1.899$ \\
    GraphSAGE-mean & $0.713$ & $0.417$ & \cellcolor{black!10}$-0.011$ & \cellcolor{black!10}$0.201$ & \cellcolor{black!10}$2.686$ \\
    H2GCN-2 & $0.884$ & $0.940$ & $0.489$ & \cellcolor{black!10}$-2.376$ & \cellcolor{black!10}$-0.396$ \\
    GPR-GNN & $0.228$ & $1.255$ & $0.742$ & \cellcolor{black!10}$0.015$ & \cellcolor{black!10}$1.545$ \\
    ACM-GCN & $0.053$ & $1.125$ & $0.755$ & \cellcolor{black!10}$0.336$ & \cellcolor{black!10}$0.424$ \\
    LINKX & $0.316$ & $0.575$ & $0.862$ & \cellcolor{black!10}$-0.010$ & $0.256$ \\
    \bottomrule
  \end{tabular*}
\end{table}

\begin{table}[H]
  \centering
  \caption{\textbf{RQ2b}. Recovered share \(\rho_{\mathrm{head}}\) after head refitting; cells are grey when the 95\% paired-bootstrap interval for the numerator contains zero or \(\hat A_{\mathrm{full}}-\hat A_{\mathrm{ERM}}<0.03\).}
  \label{tab:rho-head-recovered-share}
  \setlength{\tabcolsep}{5.5pt}
  \begin{tabular*}{\linewidth}{@{\extracolsep{\fill}}lccccc@{}}
    \toprule
    & \multicolumn{2}{c}{Low-homophily tail}
    & \multicolumn{3}{c}{High-homophily tail} \\
    \cmidrule(lr){2-3}\cmidrule(lr){4-6}
    Model & PubMed & \texttt{ogbn-arxiv} & Roman-empire & Amazon-ratings & \texttt{arxiv-year} \\
    \midrule
    GCN & $0.888$ & $0.957$ & \cellcolor{black!10}$0.421$ & \cellcolor{black!10}$-0.026$ & \cellcolor{black!10}$3.951$ \\
    GraphSAGE-mean & $0.961$ & $0.948$ & \cellcolor{black!10}$-0.032$ & \cellcolor{black!10}$-0.503$ & \cellcolor{black!10}$3.680$ \\
    H2GCN-2 & $0.751$ & $0.852$ & $0.530$ & \cellcolor{black!10}$-3.185$ & \cellcolor{black!10}$-1.657$ \\
    GPR-GNN & $0.175$ & $1.733$ & $0.997$ & \cellcolor{black!10}$-0.008$ & \cellcolor{black!10}$1.786$ \\
    ACM-GCN & $0.177$ & $0.864$ & $1.325$ & \cellcolor{black!10}$0.148$ & \cellcolor{black!10}$4.402$ \\
    LINKX & $0.719$ & $0.776$ & $1.106$ & \cellcolor{black!10}$0.029$ & $0.339$ \\
    \bottomrule
  \end{tabular*}
\end{table}

\clearpage
\subsection{RQ2: Sensitivity to Retraining Approach}
\label{app:rq5-sensitivity-retraining}

Here we present two partial ablations where we replace the GroupDRO balanced retraining loss with DFR-style Final-Layer retraining, and FG-CCDB balanced training.

\subsubsection{DFR-style Last layer Retraining}
\noindent This post-registration sensitivity analysis tests whether the RQ2a result for
a frozen representation depends on the GroupDRO optimisation used for
\(\Alast\). We repeated the retraining on PubMed for ACM-GCN, GCN, and
GraphSAGE-mean using a DFR-style retraining
\citep{kirichenko2023lastlayerretraining}. For each of the 30 matched
split--seed runs per model, we froze the ERM features supplied to the final
affine layer, standardised them, and fitted an L1-regularised multinomial
logistic regression to samples balanced across class and fixed
local-homophily-bin cells. A deterministic joint-stratified division of the
validation set was used to select the regularisation strength by worst-cell
accuracy. At the selected strength, we fitted ten classifiers to equal-cell
subsamples drawn from the full validation split and averaged their
coefficients and intercepts. We denote this procedure \(A_{\mathrm{DFR}}\).

\begin{table}[H]
  \centering
  \caption{
    PubMed tail macro-F1 for ERM, the three registered retraining procedures, and a DFR-style frozen-representation refit. Each cell gives the mean over 30 runs and the 95\% percentile-bootstrap interval.
    }
  \label{tab:rq5-dfr-tail-macro-f1}
  \setlength{\tabcolsep}{4pt}
  \begin{tabular*}{\linewidth}{@{\extracolsep{\fill}}lccccc@{}}
    \toprule
    Model & \(\Aerm\) & \(\Alast\) & \(\Ahead\) & \(\Afull\) & \(A_{\mathrm{DFR}}\) \\
    \midrule
    ACM-GCN & \shortstack{$.557$\\[-1pt]$[.550,.564]$} & \shortstack{$.562$\\[-1pt]$[.555,.569]$} & \shortstack{$.575$\\[-1pt]$[.568,.581]$} & \shortstack{$.658$\\[-1pt]$[.652,.665]$} & \shortstack{$.593$\\[-1pt]$[.586,.600]$} \\
    GCN & \shortstack{$.512$\\[-1pt]$[.506,.518]$} & \shortstack{$.537$\\[-1pt]$[.531,.543]$} & \shortstack{$.620$\\[-1pt]$[.614,.626]$} & \shortstack{$.634$\\[-1pt]$[.628,.639]$} & \shortstack{$.551$\\[-1pt]$[.542,.560]$} \\
    GraphSAGE-mean & \shortstack{$.532$\\[-1pt]$[.526,.538]$} & \shortstack{$.648$\\[-1pt]$[.635,.660]$} & \shortstack{$.688$\\[-1pt]$[.678,.697]$} & \shortstack{$.694$\\[-1pt]$[.689,.700]$} & \shortstack{$.629$\\[-1pt]$[.621,.636]$} \\
    \bottomrule
  \end{tabular*}
\end{table}

\noindent \(A_{\mathrm{DFR}}\) improves on the ERM baseline \(\Aerm\) for all three models, but is not clearly better than the GroupDRO retraining procedures. It exceeds \(\Alast\) for ACM-GCN and GCN, but not GraphSAGE-mean, and in all cases scores worse than the full model retraining of \(\Afull\). From this limited test, we conclude \(A_{\mathrm{DFR}}\) is a competitive and computationally cheaper alternative. Still, it does not qualitatively change the results from the GroupDRO analysis in the main paper.

\subsubsection{RQ2: FG-CCDB Retraining}
\label{app:fg-ccdb}

\noindent We conducted a post-registration sensitivity analysis to determine whether the RQ2 recovery gains could be reproduced by directly removing the dependence between the fixed local-homophily bin and the class label. We adapted FG-CCDB~\citep{zhao2026finegrainedccdb} by treating local-homophily bins as a discrete nuisance variable. For the eligible training nodes in each split, an observation in bin \(b\) and class \(y\) received a sampling weight.

\[
  w(b, y) = \frac{n_b n_y}{N n_{by}}.
\]

The resulting reweighted training cross-entropy batches target the product of the bin and class marginals. 

We repeated \(\Alast\), \(\Ahead\), and \(\Afull\) on PubMed for GraphSAGE-mean, GCN, and ACM-GCN, repeating the 30 training runs. Table~\ref{tab:fg-ccdb-pubmed} compares the resulting tail macro-F1 gains with those from the original GroupDRO retraining. The FG-CCDB retraining performs worse than GroupDRO across all three tested models for all retraining procedures. We conclude that FG-CCDB is not a useful alternative to our chosen GroupDRO scheme.

\begin{table}[H]
  \centering
  \caption{PubMed tail macro-F1 gain relative to \(\Aerm\) under FG-CCDB and GroupDRO. Values are means over 30 matched split--seed runs. FG-CCDB entries include 95\% percentile-bootstrap intervals; grey intervals contain zero.}
  \label{tab:fg-ccdb-pubmed}
  \begin{tabular}{@{}llcc@{}}
    \toprule
    Model & Procedure & FG-CCDB gain [95\% CI] & GroupDRO gain \\
    \midrule
    \multirow{3}{*}{GraphSAGE-mean} & \(\Alast\) & \gaincell{+.007}{[.004,.010]} & +.116 \\
     & \(\Ahead\) & \gaincell{+.043}{[.039,.048]} & +.156 \\
     & \(\Afull\) & \gaincell{+.013}{[.006,.020]} & +.162 \\
    \addlinespace[2pt]
    \multirow{3}{*}{GCN} & \(\Alast\) & \gaincell{-.015}{[-.019,-.011]} & +.025 \\
     & \(\Ahead\) & \gaincell{+.012}{[.007,.016]} & +.108 \\
     & \(\Afull\) & \uncertaingaincell{+.002}{[-.005,.008]} & +.122 \\
    \addlinespace[2pt]
    \multirow{3}{*}{ACM-GCN} & \(\Alast\) & \uncertaingaincell{-.001}{[-.004,.002]} & +.005 \\
     & \(\Ahead\) & \gaincell{+.008}{[.004,.013]} & +.018 \\
     & \(\Afull\) & \uncertaingaincell{+.003}{[-.001,.008]} & +.101 \\
    \bottomrule
  \end{tabular}
\end{table}

\section{The Effect of Edge Perturbation}
\label{app:edge-perturbation}

\noindent In this controlled experiment, we investigated the effect of changing the local homophily values of tail nodes towards the graph mean, while preserving each node's degree. We performed a graph intervention on PubMed, \texttt{ogbn-arxiv}, and \texttt{arxiv-year} using the ERM checkpoints for GCN and GraphSAGE-mean. We randomly ordered the nodes in the rare homophily tail and selected nested subsets containing \(0, 5, \ldots, 100\%\) of them. For each treatment proportion, we independently generated five treatment--control graph pairs from the original graph and report their mean results. \emph{Treatment} rewiring changed each selected node's local homophily by \(1/d_v\) towards the graph-wide mean local homophily, where \(d_v\) is its degree. The matched control performed the same number of rewiring operations without changing the selected nodes' local homophily. Both procedures preserved every node's degree. We kept the model parameters and node features fixed and evaluated 30 ERM checkpoints per model on the test split belonging to the rare-homophily tail

\begin{figure}[H]
  \centering
  \includegraphics[width=\linewidth]
  {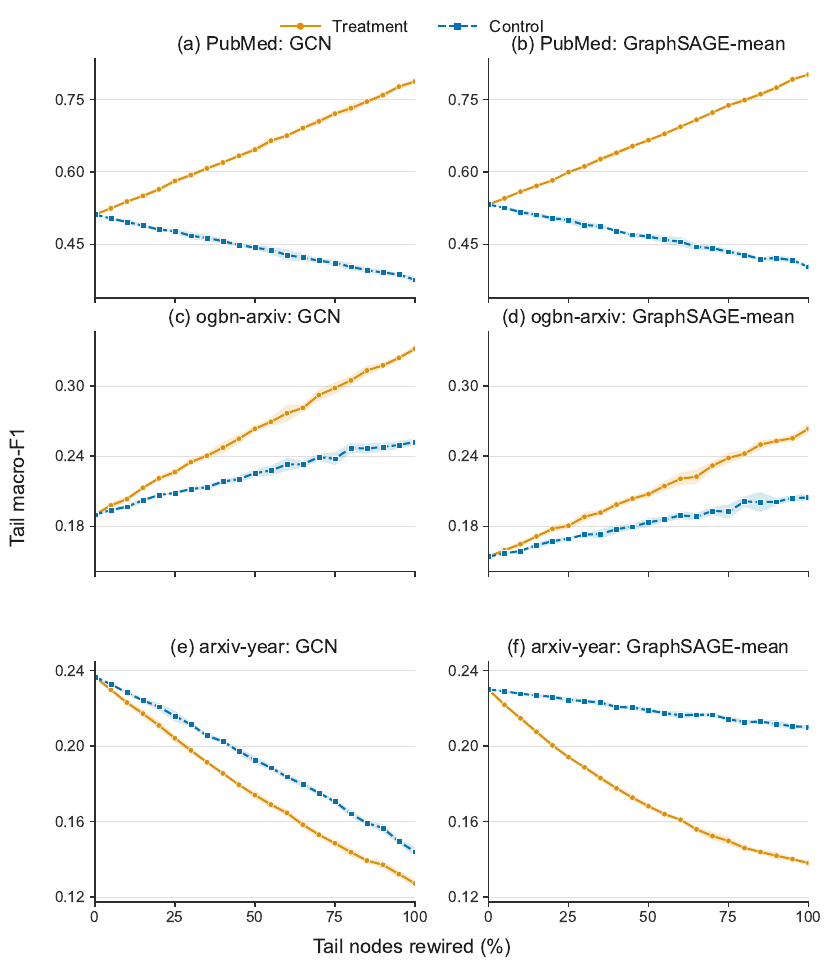}
  \caption{
    Tail macro-F1 after rewiring increasing percentages of the rare homophily tail. Treatment rewiring changes each selected node's local homophily by one neighbour towards the graph-wide mean, while the control rewires randomly, leaving the nodes' local homophily unchanged. Both procedures preserve every node's degree. Curves show means over five graph replicates after averaging over 30 fixed ERM checkpoints per model; shaded bands show one standard deviation across graph-replicate means. Treatment \emph{increases} local homophily for the homophilic datasets PubMed and \texttt{ogbn-arxiv} in (a)--(d), and decreases it for the heterophilic dataset \texttt{arxiv-year} in (e) and (f).
  }
  \label{fig:edge-perturbation-dose-response}
\end{figure}

At every non-zero percentage, treatment--control differences were positive when treatment increased local homophily (PubMed and \texttt{ogbn-arxiv}) and negative when it decreased it (\texttt{arxiv-year}). That is, making the tail more homophilic improves performance, and making a homophilic tail more heterophilic hurts performance. These results are in line with our findings in the main paper, challenging the notion that rare-homophily nodes are hard because they are rare in a graph.
\section{GenAI Usage Statement}
In this work, we used OpenAI Codex with ChatGPT 5.5 and 5.6 to assist with critiquing the research questions and experimental protocol; developing, debugging, and reviewing experimental, analysis, and plotting code; summarising relevant literature; and critiquing and revising parts of the manuscript. 

All AI-assisted material included in the paper was reviewed by the authors. AI-assisted code was inspected and tested, reported numerical results and figures were checked against saved experimental outputs, and statements about methods and prior work were checked against the project code and cited sources. 

The authors made all final methodological and interpretive decisions and take full responsibility for the paper and its accompanying artefacts.


\begin{thebibliography}{35}
\providecommand{\natexlab}[1]{#1}
\providecommand{\url}[1]{\texttt{#1}}
\expandafter\ifx\csname urlstyle\endcsname\relax
  \providecommand{\doi}[1]{doi: #1}\else
  \providecommand{\doi}{doi: \begingroup \urlstyle{rm}\Url}\fi

\bibitem[Loveland et~al.(2024)Loveland, Zhu, Heimann, Fish, Schaub, and Koutra]{loveland2023performancediscrepancieslocalhomophily}
Donald Loveland, Jiong Zhu, Mark Heimann, Benjamin Fish, Michael~T Schaub, and Danai Koutra.
\newblock On performance discrepancies across local homophily levels in graph neural networks.
\newblock In Soledad Villar and Benjamin Chamberlain, editors, \emph{Proceedings of the Second Learning on Graphs Conference}, volume 231 of \emph{Proceedings of Machine Learning Research}, pages 6:1--6:30. PMLR, 27--30 Nov 2024.
\newblock URL \url{https://proceedings.mlr.press/v231/loveland24a.html}.

\bibitem[Mao et~al.(2023)Mao, Chen, Jin, Han, Ma, Zhao, Shah, and Tang]{mao2023demystifyingstructuraldisparitygraph}
Haitao Mao, Zhikai Chen, Wei Jin, Haoyu Han, Yao Ma, Tong Zhao, Neil Shah, and Jiliang Tang.
\newblock Demystifying structural disparity in graph neural networks: can one size fit all?
\newblock In \emph{Proceedings of the 37th International Conference on Neural Information Processing Systems}, NIPS '23, Red Hook, NY, USA, 2023. Curran Associates Inc.

\bibitem[Ma et~al.(2021)Ma, Deng, and Mei]{NEURIPS2021_08425b88}
Jiaqi Ma, Junwei Deng, and Qiaozhu Mei.
\newblock Subgroup generalization and fairness of graph neural networks.
\newblock In M.~Ranzato, A.~Beygelzimer, Y.~Dauphin, P.S. Liang, and J.~Wortman Vaughan, editors, \emph{Advances in Neural Information Processing Systems}, volume~34, pages 1048--1061. Curran Associates, Inc., 2021.
\newblock URL \url{https://proceedings.neurips.cc/paper_files/paper/2021/file/08425b881bcde94a383cd258cea331be-Paper.pdf}.

\bibitem[Zhang and Dai(2026)]{zhang2026sclgnn}
Yuxiang Zhang and Enyan Dai.
\newblock Scl-gnn: Towards generalizable graph neural networks via spurious correlation learning, 2026.
\newblock URL \url{https://arxiv.org/abs/2603.08270}.

\bibitem[Sagawa et~al.(2020{\natexlab{a}})Sagawa, Koh, Hashimoto, and Liang]{sagawa2020distributionallyrobustneuralnetworks}
Shiori Sagawa, Pang~Wei Koh, Tatsunori~B. Hashimoto, and Percy Liang.
\newblock Distributionally robust neural networks for group shifts: On the importance of regularization for worst-case generalization, 2020{\natexlab{a}}.
\newblock URL \url{https://arxiv.org/abs/1911.08731}.
\newblock ICLR 2020.

\bibitem[Kirichenko et~al.(2023)Kirichenko, Izmailov, and Wilson]{kirichenko2023lastlayerretraining}
Polina Kirichenko, Pavel Izmailov, and Andrew~Gordon Wilson.
\newblock Last layer re-training is sufficient for robustness to spurious correlations, 2023.
\newblock URL \url{https://arxiv.org/abs/2204.02937}.

\bibitem[Du et~al.(2022)Du, Shi, Fu, Ma, Liu, Han, and Zhang]{du2022gbkgnngatedbikernelgraph}
Lun Du, Xiaozhou Shi, Qiang Fu, Xiaojun Ma, Hengyu Liu, Shi Han, and Dongmei Zhang.
\newblock Gbk-gnn: Gated bi-kernel graph neural networks for modeling both homophily and heterophily, 2022.
\newblock URL \url{https://arxiv.org/abs/2110.15777}.

\bibitem[Bi et~al.(2024)Bi, Du, Fu, Wang, Han, and Zhang]{bi2022makeheterophilygraphsbetter}
Wendong Bi, Lun Du, Qiang Fu, Yanlin Wang, Shi Han, and Dongmei Zhang.
\newblock { Make Heterophilic Graphs Better Fit GNN: A Graph Rewiring Approach }.
\newblock \emph{IEEE Transactions on Knowledge \& Data Engineering}, 36\penalty0 (12):\penalty0 8744--8757, December 2024.
\newblock ISSN 1558-2191.
\newblock \doi{10.1109/TKDE.2024.3441766}.
\newblock URL \url{https://doi.ieeecomputersociety.org/10.1109/TKDE.2024.3441766}.

\bibitem[Loveland and Koutra(2025)]{loveland2024unveilingimpactlocalhomophily}
Donald Loveland and Danai Koutra.
\newblock \emph{Unveiling the Impact of Local Homophily on GNN Fairness: In-Depth Analysis and New Benchmarks}, pages 608--617.
\newblock Society for Industrial and Applied Mathematics, 2025.
\newblock \doi{10.1137/1.9781611978520.65}.
\newblock URL \url{https://epubs.siam.org/doi/abs/10.1137/1.9781611978520.65}.

\bibitem[Yang et~al.(2025)Yang, Chen, Xiao, Lin, Zhang, and Kuang]{yang2025leveraginginvariantprincipleheterophilic}
Jinluan Yang, Zhengyu Chen, Teng Xiao, Yong Lin, Wenqiao Zhang, and Kun Kuang.
\newblock Leveraging invariant principle for heterophilic graph structure distribution shifts.
\newblock In \emph{Proceedings of the ACM Web Conference 2025}, pages 1196--1204. Association for Computing Machinery, 2025.
\newblock \doi{10.1145/3696410.3714749}.
\newblock URL \url{https://doi.org/10.1145/3696410.3714749}.

\bibitem[Zhu et~al.(2020)Zhu, Yan, Zhao, Heimann, Akoglu, and Koutra]{zhu2020beyondhomophilygnns}
Jiong Zhu, Yujun Yan, Lingxiao Zhao, Mark Heimann, Leman Akoglu, and Danai Koutra.
\newblock Beyond homophily in graph neural networks: current limitations and effective designs.
\newblock In \emph{Proceedings of the 34th International Conference on Neural Information Processing Systems}, NIPS '20, Red Hook, NY, USA, 2020. Curran Associates Inc.
\newblock ISBN 9781713829546.

\bibitem[Chien et~al.(2021)Chien, Peng, Li, and Milenkovic]{chien2021adaptiveuniversalgeneralizedpagerank}
Eli Chien, Jianhao Peng, Pan Li, and Olgica Milenkovic.
\newblock Adaptive universal generalized pagerank graph neural network, 2021.
\newblock URL \url{https://arxiv.org/abs/2006.07988}.
\newblock ICLR 2021.

\bibitem[Luan et~al.(2022)Luan, Hua, Lu, Zhu, Zhao, Zhang, Chang, and Precup]{luan2022revisitingheterophilygnns}
Sitao Luan, Chenqing Hua, Qincheng Lu, Jiaqi Zhu, Mingde Zhao, Shuyuan Zhang, Xiao-Wen Chang, and Doina Precup.
\newblock Revisiting heterophily for graph neural networks.
\newblock In \emph{Proceedings of the 36th International Conference on Neural Information Processing Systems}, NIPS '22, Red Hook, NY, USA, 2022. Curran Associates Inc.
\newblock ISBN 9781713871088.

\bibitem[Han et~al.(2025)Han, Li, Huang, Tang, Lu, Luo, Liu, and Tang]{han2025nodewisefilteringgraphneural}
Haoyu Han, Juanhui Li, Wei Huang, Xianfeng Tang, Hanqing Lu, Chen Luo, Hui Liu, and Jiliang Tang.
\newblock Node-wise filtering in graph neural networks: A mixture of experts approach, 2025.
\newblock URL \url{https://arxiv.org/abs/2406.03464}.

\bibitem[Lim et~al.(2021)Lim, Hohne, Li, Huang, Gupta, Bhalerao, and Lim]{lim2021largescalelearningnonhomophilous}
Derek Lim, Felix Hohne, Xiuyu Li, Sijia~Linda Huang, Vaishnavi Gupta, Omkar Bhalerao, and Ser-Nam Lim.
\newblock Large scale learning on non-homophilous graphs: new benchmarks and strong simple methods.
\newblock In \emph{Proceedings of the 35th International Conference on Neural Information Processing Systems}, NIPS '21, Red Hook, NY, USA, 2021. Curran Associates Inc.
\newblock ISBN 9781713845393.

\bibitem[Zeng et~al.(2024)Zeng, Lyu, Hu, Xia, and Luo]{zeng2024mixtureweakstrong}
Hanqing Zeng, Hanjia Lyu, Diyi Hu, Yinglong Xia, and Jiebo Luo.
\newblock Mixture of weak and strong experts on graphs, 2024.
\newblock URL \url{https://arxiv.org/abs/2311.05185}.

\bibitem[Alain and Bengio(2016)]{alain2016understanding}
Guillaume Alain and Yoshua Bengio.
\newblock Understanding intermediate layers using linear classifier probes.
\newblock \emph{arXiv preprint arXiv:1610.01644}, 2016.
\newblock URL \url{https://arxiv.org/abs/1610.01644}.

\bibitem[Belinkov(2022)]{belinkov-2022-probing}
Yonatan Belinkov.
\newblock Probing classifiers: Promises, shortcomings, and advances.
\newblock \emph{Computational Linguistics}, 48\penalty0 (1):\penalty0 207--219, March 2022.
\newblock \doi{10.1162/coli_a_00422}.
\newblock URL \url{https://aclanthology.org/2022.cl-1.7/}.

\bibitem[Akhondzadeh et~al.(2023)Akhondzadeh, Lingam, and Bojchevski]{pmlr-v206-akhondzadeh23a}
Mohammad~Sadegh Akhondzadeh, Vijay Lingam, and Aleksandar Bojchevski.
\newblock Probing graph representations.
\newblock In Francisco Ruiz, Jennifer Dy, and Jan-Willem van~de Meent, editors, \emph{Proceedings of The 26th International Conference on Artificial Intelligence and Statistics}, volume 206 of \emph{Proceedings of Machine Learning Research}, pages 11630--11649. PMLR, 25--27 Apr 2023.
\newblock URL \url{https://proceedings.mlr.press/v206/akhondzadeh23a.html}.

\bibitem[Kang et~al.(2020)Kang, Xie, Rohrbach, Yan, Gordo, Feng, and Kalantidis]{kang2019decoupling}
Bingyi Kang, Saining Xie, Marcus Rohrbach, Zhicheng Yan, Albert Gordo, Jiashi Feng, and Yannis Kalantidis.
\newblock Decoupling representation and classifier for long-tailed recognition.
\newblock In \emph{International Conference on Learning Representations}, 2020.
\newblock URL \url{https://arxiv.org/abs/1910.09217}.

\bibitem[Sagawa et~al.(2020{\natexlab{b}})Sagawa, Koh, Hashimoto, and Liang]{sagawa2020groupdrocode}
Shiori Sagawa, Pang~Wei Koh, Tatsunori~B. Hashimoto, and Percy Liang.
\newblock {group\_DRO}: Distributionally robust neural networks for group shifts, 2020{\natexlab{b}}.
\newblock URL \url{https://github.com/kohpangwei/group_DRO}.

\bibitem[Yang et~al.(2016)Yang, Cohen, and Salakhutdinov]{yang2016revisitingsemisupervisedlearninggraph}
Zhilin Yang, William~W. Cohen, and Ruslan Salakhutdinov.
\newblock Revisiting semi-supervised learning with graph embeddings.
\newblock In \emph{Proceedings of the 33rd International Conference on International Conference on Machine Learning - Volume 48}, ICML'16, page 40–48. JMLR.org, 2016.

\bibitem[Hu et~al.(2020)Hu, Fey, Zitnik, Dong, Ren, Liu, Catasta, and Leskovec]{hu2021opengraphbenchmarkdatasetsmachine}
Weihua Hu, Matthias Fey, Marinka Zitnik, Yuxiao Dong, Hongyu Ren, Bowen Liu, Michele Catasta, and Jure Leskovec.
\newblock Open graph benchmark: datasets for machine learning on graphs.
\newblock In \emph{Proceedings of the 34th International Conference on Neural Information Processing Systems}, NIPS '20, Red Hook, NY, USA, 2020. Curran Associates Inc.
\newblock ISBN 9781713829546.

\bibitem[Platonov et~al.(2024)Platonov, Kuznedelev, Diskin, Babenko, and Prokhorenkova]{platonov2024criticallookevaluationgnns}
Oleg Platonov, Denis Kuznedelev, Michael Diskin, Artem Babenko, and Liudmila Prokhorenkova.
\newblock A critical look at the evaluation of gnns under heterophily: Are we really making progress?, 2024.
\newblock URL \url{https://arxiv.org/abs/2302.11640}.

\bibitem[Pei et~al.(2020)Pei, Wei, Chang, Lei, and Yang]{pei2020geomgcn}
Hongbin Pei, Bingzhe Wei, Kevin Chen-Chuan Chang, Yu~Lei, and Bo~Yang.
\newblock Geom-gcn: Geometric graph convolutional networks, 2020.
\newblock URL \url{https://arxiv.org/abs/2002.05287}.
\newblock ICLR 2020.

\bibitem[Kipf and Welling(2017)]{kipf2017semisupervisedclassificationgraph}
Thomas~N. Kipf and Max Welling.
\newblock Semi-supervised classification with graph convolutional networks, 2017.
\newblock URL \url{https://arxiv.org/abs/1609.02907}.
\newblock ICLR 2017.

\bibitem[Hamilton et~al.(2017)Hamilton, Ying, and Leskovec]{hamilton2018inductiverepresentationlearninglarge}
William~L. Hamilton, Rex Ying, and Jure Leskovec.
\newblock Inductive representation learning on large graphs.
\newblock In \emph{Proceedings of the 31st International Conference on Neural Information Processing Systems}, NIPS'17, page 1025–1035, Red Hook, NY, USA, 2017. Curran Associates Inc.
\newblock ISBN 9781510860964.

\bibitem[Efron and Tibshirani(1994)]{efron1994introduction}
Bradley Efron and Robert~J. Tibshirani.
\newblock \emph{An Introduction to the Bootstrap}.
\newblock Chapman and Hall/CRC, New York, 1994.

\bibitem[Davison and Hinkley(1997)]{davison1997bootstrap}
A.~C. Davison and D.~V. Hinkley.
\newblock \emph{Bootstrap Methods and Their Application}.
\newblock Cambridge University Press, 1997.
\newblock \doi{10.1017/CBO9780511802843}.

\bibitem[Zhao and Zhang(2026)]{zhao2026finegrainedccdb}
Miaoyun Zhao and Qiang Zhang.
\newblock Fine-grained class-conditional distribution balancing for debiased learning, 2026.
\newblock URL \url{https://arxiv.org/abs/2505.06831}.
\newblock ICLR 2026.

\bibitem[Liu et~al.(2023)Liu, Yu, Fang, and Zhang]{10.1145/3543507.3583386}
Zemin Liu, Xingtong Yu, Yuan Fang, and Xinming Zhang.
\newblock Graphprompt: Unifying pre-training and downstream tasks for graph neural networks.
\newblock In \emph{Proceedings of the ACM Web Conference 2023}, WWW '23, page 417–428, New York, NY, USA, 2023. Association for Computing Machinery.
\newblock ISBN 9781450394161.
\newblock \doi{10.1145/3543507.3583386}.
\newblock URL \url{https://doi.org/10.1145/3543507.3583386}.

\bibitem[Yu et~al.(2025)Yu, Gong, Zhou, Fang, and Zhang]{10.1145/3696410.3714828}
Xingtong Yu, Zechuan Gong, Chang Zhou, Yuan Fang, and Hui Zhang.
\newblock Samgpt: Text-free graph foundation model for multi-domain pre-training and cross-domain adaptation.
\newblock In \emph{Proceedings of the ACM on Web Conference 2025}, WWW '25, page 1142–1153, New York, NY, USA, 2025. Association for Computing Machinery.
\newblock ISBN 9798400712746.
\newblock \doi{10.1145/3696410.3714828}.
\newblock URL \url{https://doi.org/10.1145/3696410.3714828}.

\bibitem[Fey and Lenssen(2019)]{fey2019fastgraphrepresentationlearning}
Matthias Fey and Jan~E. Lenssen.
\newblock Fast graph representation learning with pytorch geometric, 2019.
\newblock URL \url{https://arxiv.org/abs/1903.02428}.

\bibitem[Keiding and Clayton(2014)]{keiding2014standardization}
Niels Keiding and David Clayton.
\newblock Standardization and control for confounding in observational studies: A historical perspective.
\newblock \emph{Statistical Science}, 29\penalty0 (4):\penalty0 529--558, 2014.
\newblock \doi{10.1214/13-STS453}.

\bibitem[Opitz(2024)]{opitz2024closer}
Juri Opitz.
\newblock A closer look at classification evaluation metrics and a critical reflection of common evaluation practice.
\newblock \emph{Transactions of the Association for Computational Linguistics}, 12:\penalty0 820--836, 2024.
\newblock \doi{10.1162/tacl_a_00675}.
\newblock URL \url{https://aclanthology.org/2024.tacl-1.46/}.

\end{thebibliography}
\end{document}